\documentclass{article}

\usepackage{microtype}
\usepackage{graphicx}
\usepackage{subcaption}
\usepackage{booktabs} 
\usepackage{multirow}
\usepackage{paralist}
\usepackage{subcaption}
\usepackage{hyperref}
\usepackage{tikz}
\usetikzlibrary{positioning, arrows.meta, fit, calc, shapes.geometric}

\usepackage[preprint]{icml2026}

\usepackage{amsmath}
\usepackage{amssymb}
\usepackage{mathtools}
\usepackage{amsthm}
\usepackage{algorithm}

\usepackage[T1]{fontenc}
\usepackage[polish]{babel}
\usepackage[utf8]{inputenc}

\usepackage[capitalize,noabbrev]{cleveref}

\theoremstyle{plain}

\theoremstyle{definition}

\theoremstyle{remark}

\newcommand{\gflownet}{\textsc{GFlowNet}}
\newcommand{\our}{\textsc{CounterFlowNet}}

\newcommand{\PF}{P_F}

\definecolor{stage1color}{RGB}{255, 152, 0}     
\definecolor{stage2color}{RGB}{229, 57, 53}     
\definecolor{statecolor}{RGB}{227, 242, 253}    
\definecolor{stateborder}{RGB}{30, 136, 229}    
\definecolor{terminalcolor}{RGB}{67, 160, 71}   
\definecolor{arrowcolor}{RGB}{66, 66, 66}       
\definecolor{lightarrowcolor}{RGB}{150, 150, 150}       

\usepackage[textsize=tiny]{todonotes}

\icmltitlerunning{CounterFlowNet: From Minimal Changes to Meaningful Counterfactual Explanations}

\begin{document}

\twocolumn[
\icmltitle{CounterFlowNet: From Minimal Changes to Meaningful \\Counterfactual Explanations}




  \begin{icmlauthorlist}
  \icmlauthor{Oleksii Furman}{wust}
  \icmlauthor{Patryk Marszałek}{jagu,doct}
  \icmlauthor{Jan Masłowski}{jagu,doct}
  \icmlauthor{Piotr Gaiński}{jagu,doct}
  \icmlauthor{Maciej Zięba}{wust,toop}
  \icmlauthor{Marek Śmieja}{jagu}
\end{icmlauthorlist}

\icmlaffiliation{wust}{Wrocław University of Science and Technology, Wrocław, Poland}
\icmlaffiliation{jagu}{Faculty of Mathematics and Computer Science, Jagiellonian University, Kraków, Poland}
\icmlaffiliation{doct}{Doctoral School of Exact and Natural Sciences, Jagiellonian University, Kraków, Poland}
\icmlaffiliation{toop}{Tooploox Sp. z o.o., Poland}

\icmlcorrespondingauthor{Oleksii Furman}{oleksii.furman@pwr.edu.pl}

  \icmlkeywords{Counterfactual Explanations, Explainable AI, Generative Flow Networks, Machine Learning, ICML}
  \vskip 0.3in
]



\printAffiliationsAndNotice{}  

\begin{abstract}
    Counterfactual explanations (CFs) provide human-interpretable insights into model's predictions by identifying minimal changes to input features that would alter the model's output. However, existing methods struggle to generate multiple high-quality explanations that (1) affect only a small portion of the features, (2) can be applied to tabular data with heterogeneous features, and (3) are consistent with the user-defined constraints.
    We propose \our{}, a generative approach that formulates CF generation as sequential feature modification using conditional Generative Flow Networks (\gflownet{}). \our{} is trained to sample CFs proportionally to a user-specified reward function that can encode key CF desiderata: validity, sparsity, proximity and plausibility, encouraging high-quality explanations. The sequential formulation yields highly sparse edits, while a unified action space seamlessly supports continuous and categorical features. Moreover, actionability constraints, such as immutability and monotonicity of features, can be enforced at inference time via action masking, without retraining.
    Experiments on eight datasets under two evaluation protocols demonstrate that \our{} achieves superior trade-offs between validity, sparsity, plausibility, and diversity with full satisfaction of the given constraints.
\end{abstract}

\begin{figure*}[t!]
    \centering
    \includegraphics[width=0.8\textwidth]{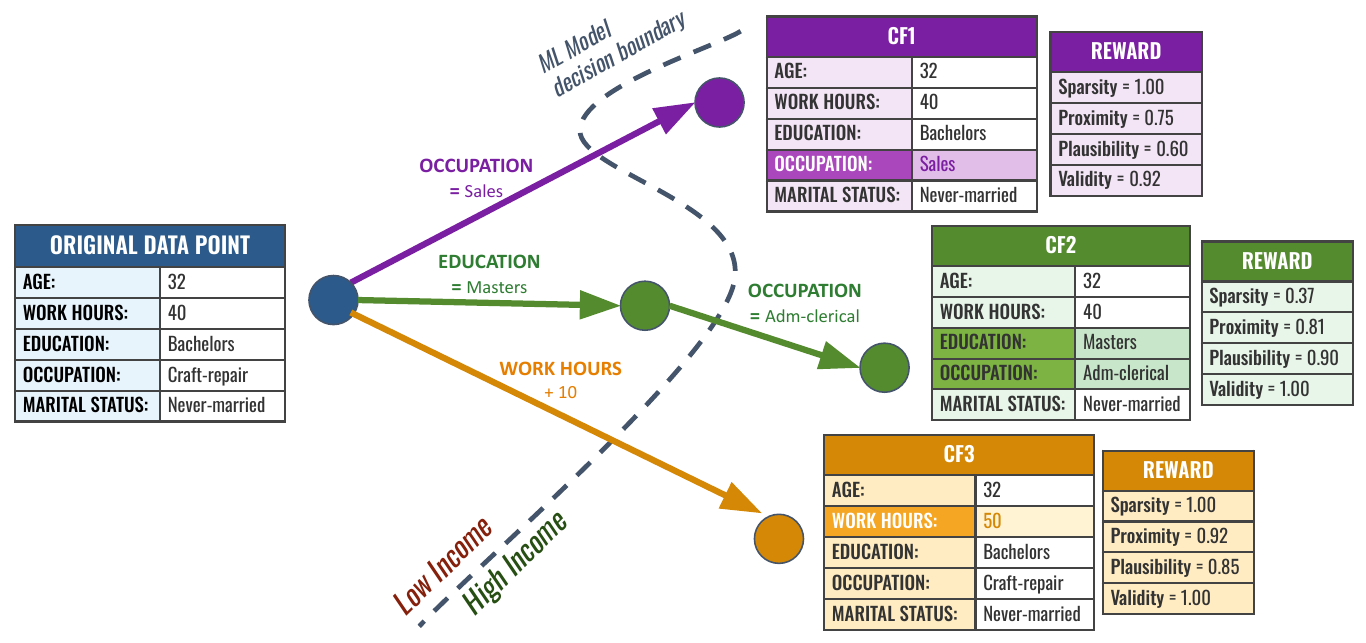}
    \caption{\textbf{\our{}} generates multiple counterfactual explanations by framing feature modification as a sequential decision process. Given an original data point (left), \our{} samples multiple valid CFs (CF1–CF3) proportional to a composite reward, naturally balancing sparsity, proximity, plausibility, and validity without requiring separate optimization.}
    \label{fig:teaser}
\end{figure*}

\section{Introduction}
\label{sec:introduction}

Counterfactual explanations (CFs) have become a central component of explainable artificial intelligence (XAI), offering human-interpretable insight into complex machine learning models \citep{guidotti2024counterfactual}. They provide actionable guidance of the form: “Your loan was denied; had your income been \$5,000 higher, you would have been approved” \citep{wachter2017counterfactual}, thereby supporting regulatory objectives such as those outlined in the GDPR \citep{goodman2017european} and forming the basis for algorithmic recourse \citep{karimi2021algorithmic, ustun2019actionable}. By proposing minimal and semantically meaningful changes to input features that would alter a model’s prediction, CFs reveal the underlying factors driving a decision and suggest feasible alternatives to achieve a desired outcome. Their value is particularly pronounced in sensitive tabular domains such as finance, healthcare, and legal decision-making, where understanding both model behavior and plausible counterfactual scenarios is critical \citep{zhou2023scgan, guidotti2019factual}.

Despite recent progress, existing methods for generating CFs still struggle to provide multiple high-quality examples offering actionable insight to the user. Optimization-based approaches such as PPCEF tend to exhibit mode-seeking behavior and converge to a single CF \citep{wielopolski2024ppcef}, while generative approaches require paired factual and counterfactual data, which are usually created based on nearest neighbors of the opposite class. \citep{pawelczyk2020learning, duong2023ceflow}. Sparse modifications are particularly desirable as they suggest actionable interventions to users, yet both generative \citep{furman2025dicoflex} and gradient-based methods \citep{wielopolski2024ppcef} frequently produce dense changes that are difficult to interpret and leverage in practice. Moreover, actionability constraints, such as feature immutability or monotonicity, remain challenging to enforce for existing methods in generation without model retraining or post-hoc filtering \citep{karimi2021algorithmic, ustun2019actionable}. Finally, real-world tabular data combines continuous and categorical features, while many existing methods assume continuous spaces \citep{wachter2017counterfactual, joshi2019towards} or require separate mechanisms for discrete attributes \citep{mothilal2020dice}.

We propose \our{}, a novel generative method that addresses these limitations through a fundamentally different formulation. Rather than optimizing a single counterfactual or a surrogate objective, \our{} uses conditional Generative Flow Networks (\gflownet{}) to sample a diverse set of high-quality examples with probability proportional to a user-defined reward. While the reward does not need to be differentiable, it can flexibly encode key counterfactual desiderata, including validity, proximity, sparsity, and plausibility. \our{} constructs explanations through sequential feature modifications, which aligns with enforcing only necessary modifications to the input example and allows us to cover discrete and continuous attributes in a unified way, see \Cref{fig:teaser}.

Our approach introduces three key design choices: (1) a \emph{two-stage state space} that decomposes each modification step into feature selection followed by value assignment, naturally accommodating both continuous and discrete features within a unified framework; (2) a \emph{composite reward function} that combines validity, proximity, sparsity, and plausibility components with tunable weights, directly incorporating quality measures into the generative training objective; and (3) \emph{constraint-aware action masking} that enforces actionability requirements directly within the policy's valid action set, eliminating the need for post-hoc filtering. Training this policy under the \gflownet{} framework produces a generative model, where the sampling probability is proportional to the reward at convergence, i.e. $P(x') \propto R(x')$, a property that justifies that the generated counterfactual examples are sampled according to the assumed quality function, with higher-quality CFs receiving a proportionally higher probability mass.

Our contributions can be summarized as follows:

\begin{compactitem}
    \item We introduce a novel generative framework for counterfactual generation, in which CFs are sampled with a probability proportional to the user-defined CFs quality reward including validity, sparsity, proximity and/or plausibility.
    \item Our framework supports mixed tabular data through a unified sequential generation process that handles both continuous and discrete attributes via a two-stage state space (feature selection + value assignment) as well as enforces sparsity of modifications.
    \item \our{} enforces actionability constraints natively through constraint-aware action masking, enabling immutability and monotonicity without post-hoc filtering or model retraining.
    \item We perform extensive experimental verification on two well-established benchmarks demonstrating a superior trade-off between diversity, sparsity, and validity compared to existing optimization-based and generative methods.
\end{compactitem}

\section{Related Work}
\label{sec:related_work}


\textbf{Counterfactual Explanations. }\citet{wachter2017counterfactual} formalized counterfactual explanations as an optimization problem balancing prediction change with proximity to the original input. \citet{mothilal2020dice} extended this framework with DiCE, introducing Determinantal Point Process regularization for diversity and one-hot encoding for categorical features. Pursuing plausibility, FACE \citep{poyiadzi2020face} searches along density-weighted $k$-NN paths, while REVISE \citep{joshi2019towards} constrains search to VAE latent spaces.  For actionability, \citet{ustun2019actionable} distinguished mutable from immutable features with monotonicity constraints, and \citet{karimi2021algorithmic}, who established theoretical foundations connecting CFs to causal interventions. However, both diversity and feature heterogeneity remain auxiliary concerns in these optimization-based frameworks. The L2C~\citep{vo2023l2c} formulates counterfactual generation as a learning problem over counting constraints, specifically designed for discretized data. Other notable approaches include MCCE \citep{redelmeier2021mcce}, which uses a Monte Carlo method to explore the manifold of valid CFs, and COPA \citep{bui2022counterfactual}, which generates CFs for structured policies and has been adapted for tabular data.


\textbf{Generative Approaches.} Generative methods amortize counterfactual generation but can be difficult to align with counterfactual quality metrics. VAE-based approaches such as C-CHVAE \citep{pawelczyk2020learning} learn class-conditional latent representations, while flow-based methods such as CeFlow \citep{duong2023ceflow} and DiCoFlex \citep{furman2025dicoflex} enable efficient sampling. However, VAE and flow-based architectures are primarily designed for continuous spaces, the relationship between sample probability and counterfactual quality depends on the training distribution rather than being explicitly specified, actionability constraints require either architectural modifications or post-hoc enforcement, and many methods require ground-truth counterfactual examples for supervised training, yet such ground truth does not exist in practice.

\textbf{Generative Flow Networks.} \gflownet~\citep{bengio2021flow} learn stochastic generators that sample objects with probability proportional to a reward, $P_\theta(x) \propto R(x)$, enabling diverse sampling from an explicitly defined quality function. They have been applied to a range of compositional domains, including biological sequences, molecular generation, and Bayesian structure learning \citep{jain2022biological, gainski2025scalable, deleu2022bayesian}. For explainability, \citet{li2023dag} proposed GFlowExplainer for diverse subgraph explanations, highlighting the potential of \gflownet{}s beyond standard generative modeling.



\begin{figure*}[t!]
\centering
\begin{tikzpicture}[
    >={Stealth[length=2.5mm]},
    state/.style={circle, draw=stateborder, fill=statecolor, line width=0.8pt, minimum size=2.5em, inner sep=1pt, font=\small},
    termstate/.style={ellipse, draw=terminalcolor, fill=terminalcolor!20, line width=0.8pt, minimum size=2.5em, inner sep=1pt, font=\small},
    ellipsestate/.style={draw=stateborder, fill=statecolor, line width=0.8pt, ellipse, minimum width=4em, minimum height=2.5em, inner sep=1pt},
    label/.style={font=\footnotesize\bfseries},
    action/.style={font=\scriptsize, color=arrowcolor},
    stage1/.style={font=\scriptsize\bfseries, color=stage1color},
    stage2/.style={font=\scriptsize\bfseries, color=stage2color},
    every node/.style={font=\scriptsize},
    node distance=3em and 4em,
    arrow/.style={->, color=arrowcolor, thick}
]


\node[ellipsestate] (s0) {$\mathbf{x}$};
\node[label, above=0.2em of s0, color=stateborder] {$s_0$};

\node[ellipsestate, right=7em of s0] (s1) {$x_{d_1} \!\leftarrow\! v_1$};
\node[label, above=0.2em of s1, color=stateborder] {$s_1$};

\draw[arrow] (s0) -- (s1) 
    node[midway, above, action] {$a_1 = (d_1, v_1)$}
    node[midway, below, action] {$\PF(a_1\!\!\mid\!\!s_0, \mathbf{x}_0, y')$};

\node[right=2em of s1, color=arrowcolor] (dots) {$\cdots$};
\draw[arrow] (s1) -- (dots);

\node[ellipsestate, right=9em of dots] (sT1) {$x_{d_{T\!-\!1}} \!\leftarrow\! v_{T\!-\!1}$};
\node[label, above=0.2em of sT1, color=stateborder] {$s_{T-1}$};

\draw[arrow] (dots) -- (sT1)
    node[midway, above, action] {$a_{T-1}$}
    node[midway, below, action] {$\PF(a_{T\!-\!1}\!\!\mid\!\!s_{T-2}, \mathbf{x}_0, y')$};

\node[termstate, right=8em of sT1] (sT) {$\mathbf{x}'$};
\node[label, above=0.2em of sT, color=terminalcolor] {$s_T$};

\draw[arrow] (sT1) -- (sT) 
    node[midway, above, action] {\textsc{stop}}
    node[midway, below, action] {$\PF(\textsc{stop}\!\!\mid\!\!s_{T-1}, \mathbf{x}_0, y')$};

\node[draw=lightarrowcolor, densely dashed, rounded corners=3pt, 
      fit=(s0)(s1), line width=1.5pt,
      inner xsep=0.5em, inner ysep=1.5em] (box) {};

\coordinate (top_center) at ($(s0)!0.5!(sT)$);

\coordinate (bottom_anchor) at ($(top_center) + (0, -8em)$);


\node[below=5em of s0, xshift=3em, state] (b_s0) {$\mathbf{x}=\begin{bmatrix} x_1 \\ \vdots \\ x_D \end{bmatrix}$};
\node[label, above=0.2em of b_s0, color=stateborder] (b_s0_label) {$s_t$};

\node[state, right=7em of b_s0, draw=stage1color, fill=stage1color!20, line width=0.8pt] (b_d) {$d_t$};

\node[state, right=8em of b_d, draw=stage2color, fill=stage2color!20, line width=0.8pt] (b_v) {$v_t$};

\node[state, right=6em of b_v] (b_s1) {$x_{d_t} \!\leftarrow\! v_t$};
\node[label, above=0.2em of b_s1, color=stateborder] (b_s1_label) {$s_{t+1}$};

\draw[->, color=stage1color, line width=1pt] (b_s0) -- (b_d) 
    node[midway, above, stage1] {\textit{Stage 1}}
    node[midway, below, action, color=stage1color] {$\PF(d\!\mid\!s_t, \mathbf{x}_0, y')$};

\draw[->, color=stage2color, line width=1pt] (b_d) -- (b_v) 
    node[midway, above, stage2] {\textit{Stage 2}}
    node[midway, below, action, color=stage2color] {$\PF(v\!\mid\!d_t, s_t, \mathbf{x}_0, y')$};

\draw[arrow] (b_v) -- (b_s1) 
    node[midway, above, action] {$a_t = (d_t, v_t)$};


\draw[color=lightarrowcolor, line width=1pt, densely dashed, shorten >=1em, shorten <=0em] 
    (box.south west) -- (b_s0_label.west);

\draw[color=lightarrowcolor, line width=1pt, densely dashed, shorten >=1em, shorten <=0em] 
    (box.south east) -- (b_s1_label.north);

\end{tikzpicture}
\caption{Overview of the \our{} generative process. 
\textbf{Top:} The forward policy $\PF$ constructs a counterfactual 
$\mathbf{x}'$ from source instance $\mathbf{x}_0$ through sequential 
feature modifications until a \textsc{stop} action terminates the trajectory. 
\textbf{Bottom:} Each action is factorized into two stages: 
sampling a feature index $d_t \sim \PF(d \mid s_t, \mathbf{x}_0, y')$ 
({\color{stage1color}\textit{Stage 1}}, Eq.~\ref{eq:index_policy}), then sampling its new value 
$v_t \sim \PF(v \mid d_t, s_t, \mathbf{x}_0, y')$ ({\color{stage2color}\textit{Stage 2}}, Eq.~\ref{eq:value_policy}).}
\label{fig:mdp_overview}
\end{figure*}
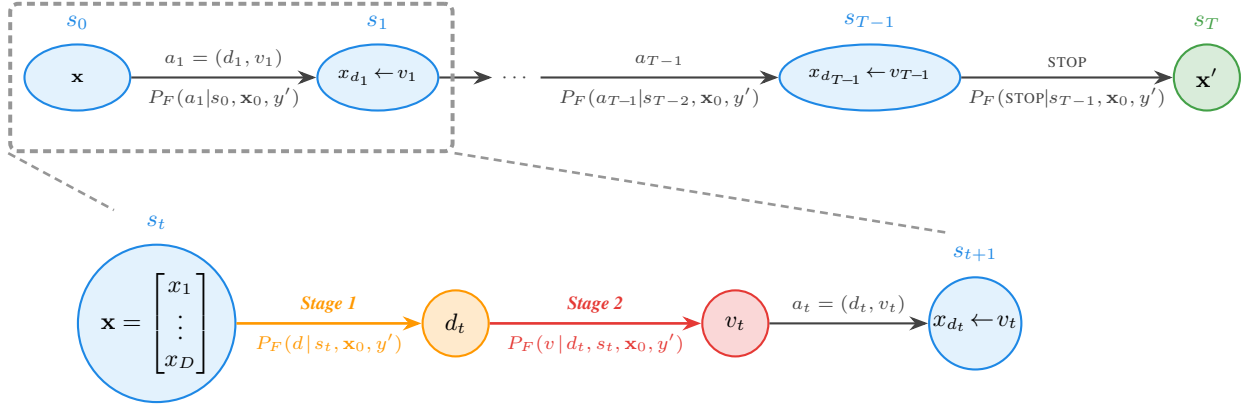

\section{Method}
\label{sec:method}
\our{} generates counterfactual explanations (CFs) by sequentially perturbing a subset of input features. At each step, it selects the feature index, followed by its perturbation. It is trained within the \gflownet{} framework, making the sampling probability proportional to the user-defined reward, encoding CF quality. In this section, we introduce the counterfactual generation problem (\Cref{sec:method_counter}) and the fundamentals of GFlowNets (\Cref{sec:method_gfns}). Next, we show how to represent the problem of probabilistic CFs with GFlowNets (\Cref{sec:method_markov}) and provide a dedicated reward function (\Cref{sec:method_reward}). Finally, we summarize training (\Cref{sec:method_trainig}) and sampling (\Cref{sec:sampling}) procedures for our approach. 

\subsection{Counterfactual Explanations Desiderata}
Given a classification model $h$ and an initial input example $\mathbf{x}_0$ within a $d$-dimensional real space $\mathbb{R}^d$, the goal is to extract counterfactual instances. A counterfactual $\mathbf{x}' \in \mathbb{R}^d$ represents a sample, which:
\begin{compactenum}
    \item[(c1)] is classified to the target class $y'$ ($y'\neq h(\mathbf{x}_0)$) by the model $h$, i.e. $h(\mathbf{x}')=y'$, \label{it1}
    \item[(c2)] lies close enough to $\mathbf{x}_0$ according to the given distance measure, $d(\mathbf{x}_0, \mathbf{x}')$, \label{it2}
    \item[(c3)] is plausible (in-distribution sample), \label{it3}
    \item[(c4)] modifies the minimal number of attributes (sparsity constraint). \label{it4}
\end{compactenum}
Typically, for many applications, there is no single best counterfactual example for a given data point satisfying the above postulates. Therefore, we are interested in modeling a conditional distribution $P(\mathbf{x}'|\mathbf{x}_0, y')$ of CFs given an example $\mathbf{x}_0$ and the target class $y'$. By sampling from $P(\mathbf{x}'|\mathbf{x}_0, y')$, we obtain diverse counterfactual options, which can be presented to the user.

\label{sec:method_counter}

\subsection{Foundations of GFlowNets}
\label{sec:method_gfns}
A \gflownet{} operates on a directed acyclic graph (DAG) $G = (\mathcal{S}, \mathbb{A})$ where: states $s \in \mathcal{S}$ represent partially constructed objects, and actions $a=(s \rightarrow s') \in \mathbb{A}$ are constructive transitions. In addition, we consider the unique source state $s_0$ and the set of terminal states $\mathcal{S}^f$. The core component of GFlowNets is the reward function $R: \mathcal{S}^f \rightarrow \mathbb{R}^{+}$ used to evaluate terminal nodes.

The goal of training a \gflownet{} is to find a non-negative flow function $F(s \rightarrow s')$ satisfying the flow-matching criterion:
\begin{align}
&\forall s: F(s)= \sum_{(s'' \rightarrow s) \in \mathbb{A}} F(s'' \rightarrow s) = \sum_{(s \rightarrow s') \in \mathbb{A}} F(s \rightarrow s'),\\
&\forall_{s_f \in \mathcal{S}^f}: \, F(s_f)=R(s_f)
\end{align}

Satisfying this criterion is equivalent to satisfying the Trajectory Balance Objective \cite{malkin2022trajectory} for every trajectory $\tau=(s_0, ..., s_T)$:
\begin{equation}
\label{eq:tb}
F(s_0)\,
\prod_{t=0}^{T-1} P_F(a_t \mid s_t)
\;=\;
R(s_f)\,
\prod_{t=0}^{T-1} P_B(a_t \mid s_{t+1}),
\end{equation}
where $P_F$ and $P_B$ are forward and backward policies. Forward policy $P_F$ defines the construction of a sample, $P_F(a_t | s_t)$ denoting the probability of taking action $a_t$ at state $s_t$, while $P_B$ governs the deconstruction analogously. $F(s_0)$ is the flow going through the source state $s_0$.

Training of the \gflownet{} converges to the policy $P_F$ that samples the terminal states in proportion to the reward:
\begin{equation}
\label{eq:p_T}
P_F(s_f)=\frac{1}{\mathcal{Z}} R(s_f) \propto R(s_f),
\end{equation}
where $P_F(s_f)$ is the probability of sampling $s_f$ following policy $P_F$, and $\mathcal{Z}$ is the normalization constant.

\subsection{Markov Decision Process of \our{}}
\label{sec:method_markov}

The generative process of \our{} is conditioned on the original input $\mathbf{x}_0$ and the target class $y'$. We adapt the \gflownet{} notation from \Cref{sec:method_gfns} by setting $s_0=\mathbf{x}_0$, allowing CFs $\mathbf{x}'$ to be the terminal state $\mathbf{x}' \in \mathcal{S}^f$, and conditioning the $P_F$, $P_B$ and $R$ with $(\mathbf{x}_0, y')$. For transparency, we omit the explicit conditioning on $y'$ if it is clear from context.

The actions for a given state represent possible feature modification decisions $a=(d,v)$, where $d$ represents index of the feature to be modified,  $d \in \{1, \ldots, D\}$, and $v \in V(d)$ is the updated feature value from a set of allowed features values for feature $d$. We can describe the process of creating counterfactual $\mathbf{x}'$ from $\mathbf{x}_0$ with the trajectory:
\begin{equation}
\tau_{\mathbf{x}_0 \to \mathbf{x}'} = (s_0=\mathbf{x}_0, s_1, \dots s_{T-1}, s_T=\mathbf{x}'),
\end{equation}
where $s_{i}$ represents the state after taking the action $a_i=(d_i,v_i)$ that set the feature $d_i$ with value $v_i$, see \Cref{fig:mdp_overview}.     

Similarly to unconditioned \gflownet{} (\Cref{eq:p_T}), properly trained $P_F$ of \our{} is in theory guaranteed to sample $x'$ proportionally to a given reward: 
\begin{equation}
P_F(\mathbf{x} ' | \mathbf{x}_0, y')\propto R(\mathbf{x}'|\mathbf{x}_0, y'),
\end{equation}
This formulation allows us to directly encode the notion of a desirable counterfactual into the generative process through the reward function $R(\mathbf{x}'|\mathbf{x}_0, y')$, which assigns higher mass to CFs that better satisfy the desired criteria (c1--c4).

Further in this section, we define our $R(\mathbf{x}'|\mathbf{x}_0, y')$, and describe parameterization and training of $P_F$.

\paragraph{Unified feature representation.}
The datasets considered in this work contain both discrete and continuous features. To enable a unified discrete-action formulation, we quantize continuous features into a finite number of bins. This allows us to flexibly model multimodal continuous distributions while avoiding unnecessary fine-grained detail. Moreover, by restricting the support of each feature to the minimum and maximum values observed during training, the conditioned \our{} operates over a discrete and countable state space, ensuring that the reward function $R(\mathbf{x}'|\mathbf{x}_0, y')$ is integrable.

\subsection{Reward function of \our{}}
\label{sec:method_reward}

The reward function $R(\mathbf{x}'|\mathbf{x}_0, y')$ of \our{} quantifies counterfactual quality by aggregating the core desiderata (c1--c4) into a single scalar signal. We adopt a multiplicative composition of four components:
\begin{multline}
\label{eq:reward_composite}
    R(\mathbf{x}'|\mathbf{x}_0, y') = R_{\text{v}}(\mathbf{x}'|y')^{\lambda_{v}} \cdot R_{\text{d}}(\mathbf{x}'| \mathbf{x}_0)^{\lambda_{d}} \\ \cdot  R_{\text{p}}(\mathbf{x}')^{\lambda_{p}} \cdot  R_{\text{s}}(\mathbf{x}'| \mathbf{x}_0)^{\lambda_{s}},
\end{multline}

where $R_{\text{v}}$ measures validity, $R_{\text{d}}$ penalizes distance (proximity), $R_{\text{p}}$ assesses plausibility, and $R_{\text{s}}$ encourages sparsity. The exponents $0\leq \lambda_v, \lambda_d, \lambda_p, \lambda_s \leq 1$ serve dual purposes: they control the relative importance of each criterion and allow users to selectively enable or disable components (setting $\lambda = 0$). The multiplicative structure ensures that \emph{all} criteria must be met, which practically means that if any factor is zero (e.g., an invalid class for validity component $R_{\text{v}}(\mathbf{x}'|y')$), the total reward is $0$. 


\textbf{Validity ($R_{\text{v}}$)} ensures the counterfactual flips the class label by penalizing dominance of the original class:
\begin{equation}
\label{eq:reward_validity}
    R_{\text{v}}(\mathbf{x}'|y') = \operatorname{clip}\Bigl(1 - \bigl(h(y_0|\mathbf{x}') -  h(y'|\mathbf{x}')\bigr) - \varepsilon,\; 0\Bigr),
\end{equation}
where $h(y'|\mathbf{x}')$ is the probability of the target class returned by the classifier $h$, $y_0$ denotes the original class label, $\varepsilon > 0$ is a margin parameter, and $\operatorname{clip}(\cdot, 0, 1)$ clamps the value to $[0,1]$. This rewards CFs where the original class is no longer dominant, with the margin $\varepsilon$ controlling the confidence threshold for the class flip.

We deliberately formulate the validity component $R_{\text{v}}$ as a continuous function, rather than using binary indicators. Although \gflownet{} rewards need not be differentiable, this continuous formulation provides graduated feedback that guides exploration toward valid CFs. A binary reward offers no signal about how close a candidate is to validity, risking inefficient exploration and suboptimal convergence.

\textbf{Proximity ($R_{\text{d}}$)} penalizes distance from the original instance:
\begin{equation}
\label{eq:reward_proximity}
    R_{\text{d}}(\mathbf{x}'|\mathbf{x}_0) = \exp\bigl(-\|\mathbf{x}' - \mathbf{x}_0\|_p\bigr),
\end{equation}
where $\|\mathbf{x}' - \mathbf{x}_0\|_p$ is the $L_p$ norm.

\textbf{Plausibility ($R_{\text{p}}$)} ensures that the counterfactual lies on the data manifold for the target class:
\begin{equation}
\label{eq:reward_plausibility}
    R_{\text{p}}(\mathbf{x}'|y') = \exp\left(\min(0, \text{LOF}(\mathbf{x}') + 1)\right),
\end{equation}
where $\text{LOF}(\mathbf{x}')$ is a Local Outlier Factor~\cite{breunig2000lof} score that compares the local density of a counterfactual point to that of its neighbors in the training set.


\textbf{Sparsity ($R_{\text{s}}$)} encourages changing as few features as possible, crucial for human interpretability:
\begin{equation}
\label{eq:reward_sparsity}
    R_{\text{s}}(\mathbf{x}'|\mathbf{x}_0) = \exp\Bigl(-\max\bigl(\|\mathbb{I}(\mathbf{x}' \neq \mathbf{x}_0)\|_0 - 1,\; 0\bigr)\Bigr),
\end{equation}
where $\mathbb{I}(\mathbf{x}' \neq \mathbf{x}_0)$ is the indicator vector for changed features and $\|\cdot\|_0$ counts the number of non-zero entries. The $-1$ offset ensures maximum reward for changing at most one feature, with exponential decay for additional changes.


\subsection{Training of \our{}}
\label{sec:method_trainig}

We train the model using the conditioned Trajectory Balance (TB) objective \cite{malkin2022trajectory} adapted from \Cref{eq:tb} to our conditioned setting:
\begin{multline}
\label{eq:tb_loss}
\mathcal{L}_{TB}(\tau) = \Bigl(\log F(\mathbf{x}_0, y') + \sum_{t=0}^{T-1} \log P_F(a_t \mid s_t, \mathbf{x}_0, y') \\
- \log R(\mathbf{x}'|\mathbf{x}_0,y') - \sum_{t=0}^{T-1} \log P_B(a_t \mid s_{t+1}, \mathbf{x}_0, y')\Bigr)^2
\end{multline}
where $P_F$ and $P_B$ denote the forward and backward policies, respectively, and $F(\mathbf{x}_0)$ is the flow associated with the initial state $s_0=\mathbf{x}_0$. Forward policy $P_F$ constructs the counterfactual by iteratively applying perturbations, while backward policy sequentially removes those perturbation, leading to original sample $\mathbf{x}_0$.

The TB objective~\eqref{eq:tb_loss} is optimized by minimizing the squared difference between the left and right sides of \Cref{eq:tb} over sampled trajectories. This formulation provably converges $P_F$ to sample CFs $\mathbf{x}'$ proportionally to their reward $R(\mathbf{x}' \mid \mathbf{x}_0, y')$. We further discuss diversity guarantees in Appendix \ref{app:diversity_guarantees}.

\textbf{Parametrization.} We set $P_B$ to be a uniform distribution and do not update it during training. We parametrize $F(\mathbf{x}_0, y')$ as a multi-layer perceptron with ReLU activations that takes as input the one-hot encoding of the original instance $\mathbf{x}_0$ and target label $y'$ and outputs a scalar flow value. 

We decompose each action $a = (d, v)$ into two sequential decisions 
(illustrated in \Cref{fig:mdp_overview}): selecting a feature index $d$, 
followed by selecting its new value $v$. The forward policy for index selection is defined as:
\begin{equation}
\label{eq:index_policy}
    P_F(d_i \mid s, \mathbf{x}_0, y') = \sigma^n(\mathbf{c})_{i},
    \quad
    \mathbf{c} = W_1 f(s),
\end{equation}
where $f : \mathcal{X} \to \mathbb{R}^D$ embeds the state $s$ using a multi-layer perceptron, and $W_1 \in \mathbb{R}^{n \times D}$ is a learnable projection matrix. The softmax operator $\sigma^k$ over a $k$-dimensional logit vector $\mathbf{c} \in \mathbb{R}^k$ is defined as:
\begin{equation}
\label{eq:softmax}
\sigma^k(\mathbf{c})_i = \frac{\exp(c_i)}{\sum_{j=1}^{k} \exp(c_j)}.
\end{equation}
At each step, we apply a mask $\mathcal{M}(s_t, \mathbf{x}_0)$ to the logits, setting $c_j = -\infty$ for indices $j$ that are either immutable by domain constraints or have already been modified in the current trajectory.

The forward policy over values $v \in V(d)$ is conditioned on the selected index $d$:
\begin{equation}
\label{eq:value_policy}
    P_F(v_i \mid d, s, \mathbf{x}_0, y') = \sigma^{|V(d)|}(\mathbf{c})_{i},
    \quad
    c_{i} = W_d f(s),
\end{equation}
where $W_d \in \mathbb{R}^{n \times D}$ is the linear embedding layer dependent on chosen feature index $d$.

\textbf{Training Procedure.} We train \our{} using a stochastic off-policy procedure detailed in \Cref{alg:training} (Appendix, \Cref{apx:training_alg}). Training proceeds by iteratively sampling batches of source instances $\mathbf{x}_0$ and target labels $y'$ from the dataset $\mathcal{D}$. For each instance, the forward policy $P_F(\cdot \mid s; \theta)$ generates a trajectory $\tau$ by sequentially modifying the state, starting from $s_0 = \mathbf{x}_0$, until termination.

At each timestep $t$, we apply a mask $\mathcal{M}(s_t, \mathbf{x}_0)$ to the policy logits, enforcing domain constraints and preventing re-modification of features within the same trajectory. Actions are sampled in two stages: first selecting a feature index $d$, then a value $v$ from the valid domain $V(d)$. Upon reaching a terminal state $\mathbf{x}'$, we compute the composite reward $R(\mathbf{x}' \mid \mathbf{x}_0, y')$ (\Cref{eq:reward_composite}). The policy parameters $\theta$ and flow estimator parameters $\psi$ are updated jointly via gradient descent on $\mathcal{L}_{\text{TB}}$ (\Cref{eq:tb_loss}).

\subsection{Sampling Counterfactual Explanations}
\label{sec:sampling}

Once trained, the forward policy $P_F$ enables efficient, amortized generation of CFs without iterative optimization. Given an input instance $\mathbf{x}_0$ and target class $y'$, we generate multiple candidates by sampling $K$ independent trajectories from $P_F$.

\textbf{Trajectory rollout.} Each trajectory $\tau$ begins at the initial state $s_0 = \mathbf{x}_0$. At step $t$, we compute an action mask $m_t = \mathcal{M}(s_t, \mathbf{x}_0)$, then sample a feature index $d_t$ and new value $v_t$ from the masked policy distributions (\Cref{eq:index_policy,eq:value_policy}). The process terminates when the policy selects the \textsc{STOP} action or a maximum sparsity budget is reached. The final state $s_T$ represents the generated CF $\mathbf{x}'$.

\textbf{Constraints handling.} We enforce actionability constraints directly during sampling via constraint-aware action masking. The mask $m_t$ removes invalid actions from the forward policy’s support, including (i) modifications of immutable features, (ii) actions that would violate monotonicity constraints (e.g., decreasing a feature that must be non-decreasing), and (iii) re-modifying a feature already changed earlier in the same trajectory. Since sampling is performed from the renormalized masked distributions, every generated trajectory respects the specified constraints without retraining or post-hoc filtering.

\textbf{Continuous Feature Reconstruction.}
\our{} operates on a discretized state space where actions correspond to bin selections. To map a discrete state $\mathbf{x}'_{\text{disc}}$ back to the continuous domain, we apply bin-center reconstruction:
\begin{equation}
\label{eq:decode}
    x'_{j} = \ell_j + (b_j + 0.5) \cdot \frac{u_j - \ell_j}{B},
\end{equation}
where $b_j \in \{0, \dots, B-1\}$ is the bin index for feature $j$, and $[\ell_j, u_j]$ denotes the feature range.

\textbf{Candidate Selection.}
Since $P_F$ samples proportionally to the reward, the batch of $K$ trajectories naturally covers high-quality modes of the counterfactual distribution. We filter the candidates to retain only valid CFs satisfying $h(\mathbf{x}') = y'$, then select the top-$N$ unique instances ranked by $R(\mathbf{x}' \mid \mathbf{x}_0, y')$.

\begin{table}[t]
\caption{Results on Protocol A datasets (discretized features, $B=4$).
Best in \textbf{bold}, second-best \underline{underlined}.}
\label{tab:protocol_A}
\centering
\tiny
\setlength{\tabcolsep}{3pt}
\resizebox{\columnwidth}{!}{%
\begin{tabular}{lcccccc}
\toprule
\textbf{Method} & \textbf{Spars.(\%)$\uparrow$} & \textbf{Div.(\%)$\uparrow$} & \textbf{H. Mean(\%)$\uparrow$} & \textbf{Val.(\%)$\uparrow$} & \textbf{Cov.(\%)$\uparrow$} & \textbf{Unary(\%)$\uparrow$} \\
\midrule
\multicolumn{7}{c}{\textit{German Credit (Logistic Regression)}} \\
\midrule
L2C & 61.35 & 37.31 & \underline{46.39} & \textbf{100.00} & \textbf{100.00} & \underline{99.06} \\
DICE-R & \textbf{88.23} & 15.29 & 26.06 & \textbf{100.00} & \textbf{100.00} & 90.81 \\
DICE-G & 43.45 & \underline{37.56} & 40.29 & 62.87 & \underline{90.24} & 56.66 \\
COPA & 57.88 & 18.88 & 28.47 & 44.00 & 44.00 & 84.31 \\
MCCE & 28.76 & 33.40 & 30.91 & 48.74 & \textbf{100.00} & 58.76 \\
CFN & \underline{69.09} & \textbf{41.26} & \textbf{51.65} & \underline{99.48} & \textbf{100.00} & \textbf{100.00} \\
\midrule
\multicolumn{7}{c}{\textit{Adult Income (Neural Network)}} \\
\midrule
L2C & 45.70 & \underline{28.11} & \underline{34.80} & \textbf{100.00} & \textbf{100.00} & \underline{97.62} \\
DICE-R & \textbf{89.26} & 9.05 & 16.44 & \textbf{100.00} & \textbf{100.00} & 87.15 \\
DICE-G & 41.48 & 26.27 & 32.14 & \underline{92.64} & \textbf{100.00} & 72.70 \\
MCCE & 24.93 & 4.58 & 7.74 & 30.63 & \underline{74.76} & 45.79 \\
CFN & \underline{62.36} & \textbf{38.21} & \textbf{47.39} & \textbf{100.00} & \textbf{100.00} & \textbf{100.00} \\
\midrule
\multicolumn{7}{c}{\textit{Graduate Admission (Neural Network)}} \\
\midrule
L2C & 42.23 & 37.90 & 39.94 & \textbf{100.00} & \textbf{100.00} & \textbf{100.00} \\
DICE-R & \textbf{66.25} & 30.93 & \underline{42.15} & \textbf{100.00} & \textbf{100.00} & \underline{85.30} \\
DICE-G & 23.05 & \textbf{47.54} & 31.04 & 92.91 & \textbf{100.00} & 66.69 \\
MCCE & 17.39 & 22.98 & 19.51 & 43.79 & \underline{84.60} & 79.11 \\
CFN & \underline{55.41} & \underline{46.18} & \textbf{50.37} & \underline{99.53} & \textbf{100.00} & \textbf{100.00} \\
\midrule
\multicolumn{7}{c}{\textit{Student Performance (Logistic Regression)}} \\
\midrule
L2C & 55.32 & 29.54 & 38.51 & \textbf{100.00} & \textbf{100.00} & \textbf{100.00} \\
DICE-R & \textbf{87.60} & 13.64 & 23.60 & \textbf{100.00} & \textbf{100.00} & \underline{98.99} \\
DICE-G & 39.20 & \textbf{39.88} & \underline{38.54} & 84.83 & \textbf{100.00} & 60.77 \\
COPA & 50.45 & 25.28 & 33.68 & 67.26 & 67.26 & 95.32 \\
MCCE & 25.97 & 24.97 & 25.46 & 60.98 & \underline{93.10} & 67.70 \\
CFN & \underline{71.18} & \underline{31.44} & \textbf{43.61} & \underline{99.63} & \textbf{100.00} & \textbf{100.00} \\
\bottomrule
\end{tabular}
}
\end{table}

\section{Experiments}
\label{sec:experiments}

\begin{table}[t]
\caption{Results on Protocol B datasets (discretized features, $B=64$).
Best in \textbf{bold}, second-best \underline{underlined}.}
\label{tab:protocol_B}
\centering
\tiny
\setlength{\tabcolsep}{3pt}
\resizebox{\columnwidth}{!}{%
\begin{tabular}{lcccccc}
\toprule
\textbf{Method} & \textbf{Val. $\uparrow$} & \textbf{Prox.-Cont $\downarrow$} & \textbf{Spars.-Cat $\downarrow$} & \textbf{$\epsilon$-Spars. $\downarrow$} & \textbf{LOF $\downarrow$} & \textbf{Div. $\uparrow$} \\
\midrule
\multicolumn{7}{c}{\textit{Adult Income  (Neural Network)}} \\
\midrule
DICE  & \textbf{1.00} & \underline{0.44} & \textbf{0.05} & \underline{0.50} & \textbf{0.17} & 0.06 \\
CCHVAE & \textbf{1.00} & 0.50 & \underline{0.06} & 0.98 & 0.22 & 0.03 \\
DiCoFlex & \textbf{1.00} & 0.85 & 0.49 & 0.97 & 0.31 & \textbf{0.40} \\
CFN & \textbf{1.00} & \textbf{0.19} & 0.27 & \textbf{0.36} & \underline{0.20} & \underline{0.21} \\
\midrule
\multicolumn{7}{c}{\textit{Bank (Neural Network)}} \\
\midrule
DICE  & \textbf{1.00} & \underline{0.61} & \textbf{0.07} & \underline{0.66} & \textbf{0.11} & 0.12 \\
CCHVAE & \textbf{1.00} & 0.73 & \underline{0.15} & 0.97 & \underline{0.15} & 0.08 \\
DiCoFlex & \textbf{1.00} & 0.89 & 0.40 & 0.95 & 0.16 & \textbf{0.39} \\
CFN & \textbf{1.00} & \textbf{0.25} & 0.23 & \textbf{0.38} & 0.21 & \underline{0.21} \\
\midrule
\multicolumn{7}{c}{\textit{Default (Neural Network)}} \\
\midrule
DICE  & \textbf{1.00} & \underline{0.31} & \textbf{0.06} & \underline{0.77} & \textbf{0.24} & 0.05 \\
CCHVAE & \textbf{1.00} & 0.48 & \underline{0.10} & 0.97 & \underline{0.33} & 0.04 \\
DiCoFlex & \textbf{1.00} & 0.64 & 0.56 & 0.97 & 0.39 & \textbf{0.33} \\
CFN & \underline{0.96} & \textbf{0.12} & 0.26 & \textbf{0.19} & 0.34 & \underline{0.14} \\
\midrule
\multicolumn{7}{c}{\textit{GMC (Neural Network)}} \\
\midrule
DICE  & \textbf{1.00} & \underline{0.26} & \textbf{0.04} & \underline{0.68} & \textbf{0.05} & 0.04 \\
CCHVAE & \textbf{1.00} & 0.43 & \underline{0.07} & 0.96 & 0.57 & 0.01 \\
DiCoFlex & \textbf{1.00} & 0.79 & 0.83 & 0.96 & 0.53 & \textbf{0.51} \\
CFN & \textbf{1.00} & \textbf{0.12} & 0.63 & \textbf{0.25} & \underline{0.24} & \underline{0.18} \\
\midrule
\multicolumn{7}{c}{\textit{Lending Club (Neural Network)}} \\
\midrule
DICE  & \textbf{1.00} & 0.82 & \textbf{0.19} & \underline{0.79} & \textbf{0.03} & 0.13 \\
CCHVAE & \textbf{1.00} & \underline{0.63} & 0.65 & 0.94 & 0.15 & 0.05 \\
DiCoFlex & \textbf{1.00} & 1.06 & 0.76 & 0.94 & \underline{0.05} & \textbf{0.34} \\
CFN & \underline{0.94} & \textbf{0.56} & \underline{0.41} & \textbf{0.29} & 0.21 & \underline{0.24} \\
\bottomrule
\end{tabular}
}
\end{table}

We evaluate \our{} on eight benchmark datasets under two evaluation protocols, demonstrating state-of-the-art validity-diversity trade-offs across varying feature representations. Additional details on our experimental setup and implementation are available in Appendix~\ref{sec:implementation}. Furthermore, we defer complementary analyses to the appendix, including a feature-utilization study (Appendix~\ref{app:feature_utilization}) and qualitative case studies with representative counterfactual examples (Appendix~\ref{app:qualitative_examples}).

\subsection{Experimental Setup}
\textbf{Datasets and Evaluation Protocols.}
We adopt datasets from two recent counterfactual benchmarks that
employ complementary evaluation philosophies.
\textbf{Protocol A} assumes discrete features for all baseline methods \citep{vo2023l2c}. Continuous features are quantized into equal-frequency bins $B=4$ prior to classifier training.
\textbf{Protocol B} evaluates methods using continuous and discrete features \citep{furman2025dicoflex}. Baseline methods are trained on the original mixed continuous-categorical space.

\textbf{Bridging the Protocols.} \our{} operates on discrete action spaces by design. For Protocol B, we quantize continuous features into $B=64$ bins at the \emph{counterfactual generation} stage while preserving the original classifier, then decode via bin-center reconstruction (Eq.~\ref{eq:decode}). This enables tractable sequential generation, provides a tunable proximity-diversity trade-off via $B$, and matches the implicit discretization in competitor methods (e.g., DiCoFlex's $L_p$ neighbor selection operates on finite training examples). Dataset characteristics are detailed in Appendix~\ref{app:datasets}.

\begin{figure}
    \centering
    \includegraphics[width=\linewidth]{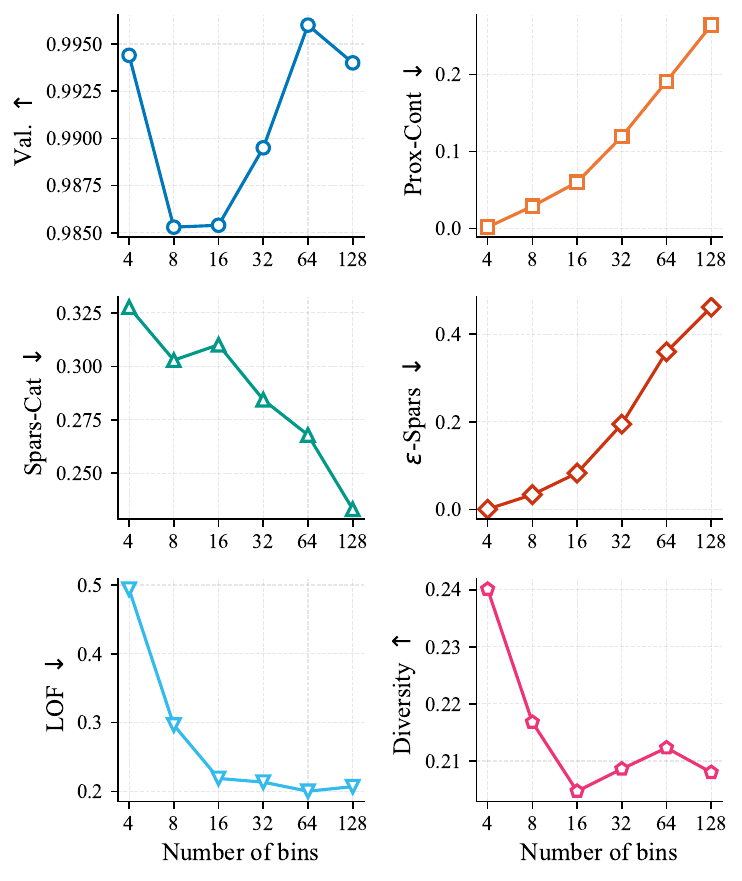}
    \caption{Effect of discretization granularity $B$ on Adult dataset.}
    \label{fig:discretization_ablation}
\end{figure}

\textbf{Baselines.}
We compare \our{} against a comprehensive set of state-of-the-art methods, spanning three major paradigms: optimization-based (DiCE, COPA), sampling-based (MCCE), and generative/amortized methods (L2C, C-CHVAE, DiCoFlex). A detailed description of each baseline is provided in Appendix~\ref{app:baselines}.

\textbf{Evaluation Metrics}
We evaluate validity, proximity, sparsity, diversity, plausibility (LOF), and constraint satisfaction, following the metric definitions of \citet{vo2023l2c} for Protocol A and \citet{furman2025dicoflex} for Protocol B. Detailed formulations are in Appendix~\ref{app:metrics}.







\begin{figure*}[t]
  \centering
  \begin{subfigure}[b]{0.32\textwidth}
    \centering
    \includegraphics[width=\textwidth]{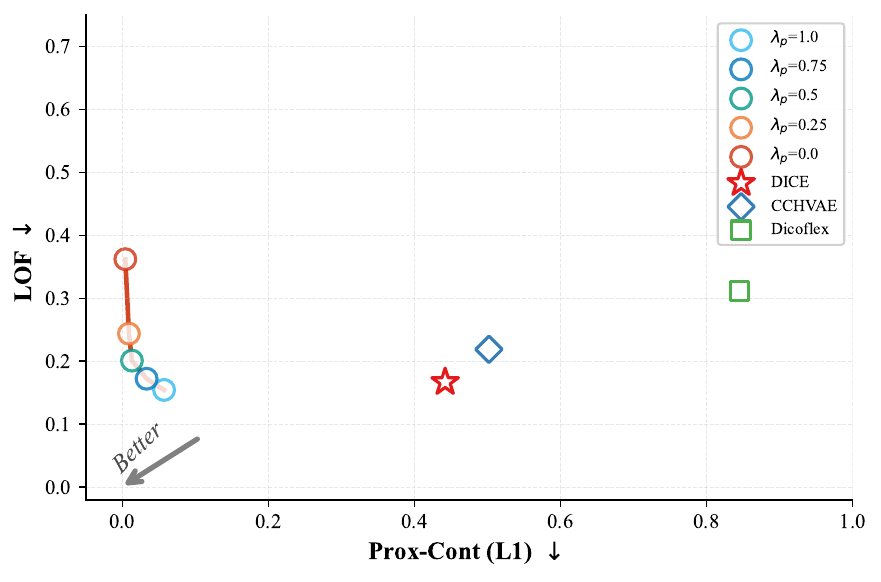}
    \caption{Varying $\lambda_p$ (plausibility weight) with $\lambda_d{=}1.0$, $\lambda_v{=}1.0$, $\lambda_s{=}0.4$ fixed.}
    \label{fig:ablation-a}
  \end{subfigure}
  \hfill
  \begin{subfigure}[b]{0.32\textwidth}
    \centering
    \includegraphics[width=\textwidth]{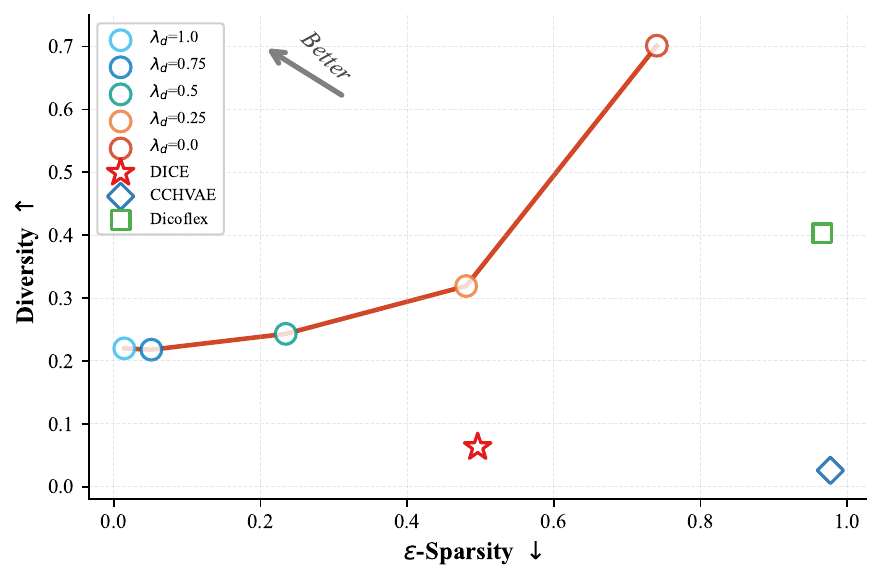}
    \caption{Varying $\lambda_d$ (proximity weight) with $\lambda_p{=}0.0$, $\lambda_v{=}1.0$, $\lambda_s{=}0.4$ fixed.}
    \label{fig:ablation-c}
  \end{subfigure}
  \hfill
  \begin{subfigure}[b]{0.32\textwidth}
    \centering
    \includegraphics[width=\textwidth]{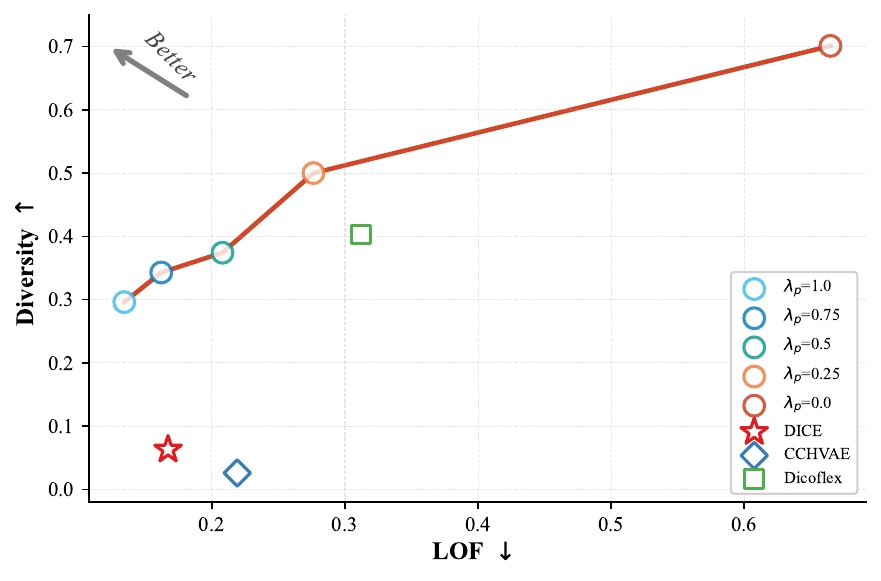}
    \caption{Varying $\lambda_d$ (proximity weight) with $\lambda_p{=}1.0$, $\lambda_v{=}1.0$, $\lambda_s{=}0.4$ fixed.}
    \label{fig:ablation-b}
  \end{subfigure}
  \caption{Reward component ablation on the Adult dataset. (a) Increasing plausibility weight $\lambda_p$ improves LOF at the cost of proximity. (b-c) Higher proximity weight $\lambda_d$ yields sparser, more plausible CFs with lower diversity. All \our{} configurations consistently dominate the baseline Pareto frontiers.}
  \label{fig:ablation_weights}
\end{figure*}

\subsection{Main Results}
\label{sec:results}



\textbf{Protocol A (Discretized).} Table~\ref{tab:protocol_A} compares methods on the L2C benchmark
where all features are discretized ($B=4$). This setting directly
matches \our{}'s discrete action space. \our{} demonstrates a superior balance between all key desiderata. As shown in Table~\ref{tab:protocol_A}, our method consistently achieves near-perfect Validity, Coverage, and Unary constraint satisfaction ($\geq$99\%), outperforming most baselines, particularly DICE-G, COPA, and MCCE, which often fail to generate valid or constraint-abiding CFs. While DICE-R achieves high sparsity, it does so at the expense of diversity. Conversely, methods like DICE-G and L2C achieve higher diversity but compromise on sparsity. \our{} achieves the highest Harmonic Mean across all four datasets, demonstrating superior balance between these competing objectives.

\textbf{Protocol B (Continuous).} In the continuous protocol, we compare \our{} with three state-of-the-art methods adapted to this setting, namely DiCE (KD-Tree), CCHVAE, and DiCoFlex. In the continuous setting (Table~\ref{tab:protocol_B}), \our{} demonstrates state-of-the-art performance in generating sparse and close CFs while maintaining high diversity. It consistently achieves the best proximity (Prox-Cont) and $\epsilon$-sparsity, indicating that its modifications are both small in magnitude and few in number. While DiCE and CCHVAE produce slightly sparser changes for categorical features on some datasets, they fall short on numerical sparsity. DiCoFlex achieves the highest diversity, but our method offers a much better balance, ranking second in diversity while being significantly more sparse and proximal. All methods generate valid and plausible (low LOF) CFs, but \our{} provides effective trade-off between minimal, interpretable changes and diverse recourse options.

\subsection{Ablation Studies}
\label{sec:ablations}

\textbf{Discretization Granularity.} Figure~\ref{fig:discretization_ablation} analyzes sensitivity to the number of bins $B$ on the Adult dataset. As $B$ increases, the finer-grained action space yields more plausible CFs closer to the data manifold (decreasing LOF), but at the cost of worse proximity and $\epsilon$-sparsity due to accumulated small adjustments. Diversity peaks at lower $B$, suggesting that coarser grids force exploration of structurally different solutions. Validity remains high in all settings, reaching its peak at $B=64$. Our choice of $B=64$ represents a balanced compromise between plausibility and other desiderata.

\textbf{Impact of Constraints.} We evaluate how \our{} handles user-specified actionability requirements by progressively tightening constraints on the Adult dataset. \textit{Set 1} fixes \textit{sex} and \textit{race} and enforces monotonicity on \textit{age}; \textit{Set 2} additionally makes \textit{native country} immutable and enforces monotonicity on \textit{education}; \textit{Set 3} further fixes \textit{marital status}. For each set, we compare two ways of enforcing constraints: (i) \emph{Inference-time} masking, where we apply constraint-aware action masking $\mathcal{M}(s_t,\mathbf{x}_0)$ during sampling without retraining, and (ii) \emph{retraining} with constraints, where the same mask is used throughout training. Across constraint sets, tightening actionability requirements generally degrades proximity and sparsity gracefully, while validity remains high, highlighting a clear trade-off between feasibility constraints and minimality of edits. Overall, inference-time masking achieves performance comparable to retraining while guaranteeing 100\% constraint satisfaction by construction, making it a practical way to deploy \our{} under new user constraints without additional training.

\textbf{Reward Components.} The composite reward allows practitioners to balance competing desiderata via weights $\lambda_v, \lambda_d, \lambda_p, \lambda_s$. Figure~\ref{fig:ablation_weights} illustrates these trade-offs. Increasing the plausibility weight $\lambda_p$ improves LOF at the cost of proximity (Fig.~\ref{fig:ablation-a}). Higher proximity weight $\lambda_d$ produces sparser and more plausible CFs with reduced diversity (Fig.~\ref{fig:ablation-b},~\ref{fig:ablation-c}). Crucially, all \our{} configurations consistently define a better Pareto frontier than DiCE and C-CHVAE, demonstrating that the framework provides fine-grained control while maintaining competitive performance. Full quantitative results are in Appendix~\ref{app:reward_ablation}.

\begin{table}[t]
\caption{Impact of actionability constraints on Adult. CFs quality under three increasingly strict constraint sets, comparing \emph{inference-time} constraint action masking to \emph{retraining} with the same constraints applied during training.}
\label{tab:ablations_constraints}
\centering
\tiny
\setlength{\tabcolsep}{3pt}
\resizebox{\columnwidth}{!}{%
\begin{tabular}{lcccccc}
\toprule
\textbf{CFN} & \textbf{Val. $\uparrow$} & \textbf{Prox.-Cont $\downarrow$} & \textbf{Spars.-Cat $\downarrow$} & \textbf{$\epsilon$-Spars. $\downarrow$} & \textbf{LOF $\downarrow$} & \textbf{Div. $\uparrow$} \\
\midrule
\multicolumn{7}{c}{\textit{Adult Income - Unconstrained}} \\
\midrule
Retrained & 0.99 & 0.22 & 0.28 & 0.41 & 0.22 & 0.23 \\
\midrule
\multicolumn{7}{c}{\textit{Adult Income - Constraints Set 1}} \\
\midrule
Retrained & 0.99 & 0.24 & 0.27 &0.45 & 0.22 & 0.23\\
Inference & 0.99 & 0.24 & 0.27 & 0.46 & 0.22 & 0.23 \\
\midrule
\multicolumn{7}{c}{\textit{Adult Income - Constraints Set 2}} \\
\midrule
Retrained & 0.98 & 0.24 & 0.28 & 0.44 & 0.22 & 0.21 \\
Inference & 0.99 & 0.25 & 0.26 & 0.48 & 0.23 & 0.20 \\
\midrule
\multicolumn{7}{c}{\textit{Adult Income - Constraints Set 3}} \\
\midrule
Retrained & 0.99 & 0.33 & 0.25 & 0.49 & 0.24 & 0.21 \\
Inference & 0.97 & 0.28 & 0.25 & 0.50 & 0.25 & 0.19 \\
\bottomrule
\end{tabular}
}
\end{table}

\textbf{Additional results.} Due to space limitations, we include two complementary analyses in the appendix. Appendix~\ref{app:feature_utilization} provides a comparative feature-utilization study, showing which features are most frequently modified by \our{} and the baselines. Appendix~\ref{app:qualitative_examples} provides qualitative case studies, illustrating representative CFs and the types of changes suggested by our method.

\section{Conclusions}

We introduce \our{}, a method that frames CFs generation as a sequential decision process and learns to sample CFs with probability proportional to a user-defined reward, directly encoding validity, proximity, sparsity, and plausibility. \our{} supports mixed data types and enforces actionability constraints via inference-time action masking. Across eight datasets and two evaluation protocols, \our{} achieves a superior trade-off between validity, sparsity, plausibility, and diversity compared to state-of-the-art baselines. Future work includes richer reward functions (e.g., fairness), extending to new modalities such as text, and multi-objective \gflownet{}s for navigating trade-offs.



\section*{Impact Statement}
This work introduces \our{}, a method for generating counterfactual explanations to improve the transparency and interpretability of machine learning models. The primary societal benefit of our work is the potential to empower individuals subject to algorithmic decisions. By providing diverse, sparse, and actionable recourse, \our{} can help users understand why a particular decision was made and what steps they can take to achieve a different outcome. This can foster greater trust and accountability, particularly in high-stakes domains like finance, healthcare, and employment. Furthermore, by allowing for the inspection of model behavior under constraints, our method can serve as a tool for developers and auditors to identify and mitigate potential model biases.

\bibliography{main}

@article{zhou2023scgan,
  title={SCGAN: Sparse CounterGAN for counterfactual explanations in breast cancer prediction},
  author={Zhou, Siqiong and Islam, Upala J and Pfeiffer, Nicholaus and Banerjee, Imon and Patel, Bhavika K and Iquebal, Ashif S},
  journal={IEEE Transactions on Automation Science and Engineering},
  volume={21},
  number={3},
  pages={2264--2275},
  year={2023},
  publisher={IEEE}
}

@article{koziarski2024rgfn,
  title={{RGFN}: Synthesizable molecular generation using {GFlowNets}},
  author={Koziarski, Micha{\l} and Rekesh, Andrei and Shevchuk, Dmytro and van der Sloot, Almer and Gai{\'n}ski, Piotr and Bengio, Yoshua and Liu, Chenghao and Tyers, Mike and Batey, Robert},
  journal={Advances in Neural Information Processing Systems},
  volume={37},
  pages={46908--46955},
  year={2024}
}

@inproceedings{
gainski2025scalable,
title={Scalable and Cost-Efficient de Novo Template-Based Molecular Generation},
author={Piotr Gai{\'n}ski and Oussama Boussif and Andrei Rekesh and Dmytro Shevchuk and Ali Parviz and Mike Tyers and Robert A. Batey and Micha{\l} Koziarski},
booktitle={The Thirty-ninth Annual Conference on Neural Information Processing Systems},
year={2025},
url={https://openreview.net/forum?id=zssWxiiJZ1}
}

@inproceedings{breunig2000lof,
  title={LOF: identifying density-based local outliers},
  author={Breunig, Markus M and Kriegel, Hans-Peter and Ng, Raymond T and Sander, J{\"o}rg},
  booktitle={Proceedings of the 2000 ACM SIGMOD international conference on Management of data},
  pages={93--104},
  year={2000}
}

@inproceedings{wielopolski2024ppcef,
  author       = {Patryk Wielopolski and
                  Oleksii Furman and
                  Jerzy Stefanowski and
                  Maciej Zieba},
  title        = {Probabilistically Plausible Counterfactual Explanations with Normalizing
                  Flows},
  booktitle    = {{ECAI} 2024 - 27th European Conference on Artificial Intelligence},
  volume       = {392},
  pages        = {954--961},
  publisher    = {{IOS} Press},
  year         = {2024},
  url          = {https://doi.org/10.3233/FAIA240584},
  doi          = {10.3233/FAIA240584},
}

@article{guidotti2019factual,
  title={Factual and counterfactual explanations for black box decision making},
  author={Guidotti, Riccardo and Monreale, Anna and Giannotti, Fosca and Pedreschi, Dino and Ruggieri, Salvatore and Turini, Franco},
  journal={IEEE Intelligent Systems},
  volume={34},
  number={6},
  pages={14--23},
  year={2019},
  publisher={IEEE}
}

@article{guidotti2024counterfactual,
  title={Counterfactual explanations and how to find them: literature review and benchmarking},
  author={Guidotti, Riccardo},
  journal={Data Mining and Knowledge Discovery},
  volume={38},
  number={5},
  pages={2770--2824},
  year={2024},
  publisher={Springer}
}

@article{wachter2017counterfactual,
  title={Counterfactual Explanations without Opening the Black Box: Automated Decisions and the {GDPR}},
  author={Wachter, Sandra and Mittelstadt, Brent and Russell, Chris},
  journal={Harvard Journal of Law \& Technology},
  volume={31},
  number={2},
  pages={841--887},
  year={2017}
}

@inproceedings{mothilal2020dice,
  title={Explaining Machine Learning Classifiers through Diverse Counterfactual Explanations},
  author={Mothilal, Ramaravind K and Sharma, Amit and Tan, Chenhao},
  booktitle={Proceedings of the 2020 Conference on Fairness, Accountability, and Transparency},
  pages={607--617},
  year={2020},
  organization={ACM}
}

@inproceedings{poyiadzi2020face,
  title={{FACE}: Feasible and Actionable Counterfactual Explanations},
  author={Poyiadzi, Rafael and Sokol, Kacper and Santos-Rodriguez, Raul and De Bie, Tijl and Flach, Peter},
  booktitle={Proceedings of the AAAI/ACM Conference on AI, Ethics, and Society},
  pages={344--350},
  year={2020}
}

@inproceedings{vo2023l2c,
  title={Learning to Counter: Stochastic Feature-based Learning for Diverse Counterfactual Explanations},
  author={Vo, Vy and Le, Trung and Nguyen, Van and Zhao, He and Bonilla, Edwin and Haffari, Gholamreza and Phung, Dinh},
  booktitle={Proceedings of the 29th ACM SIGKDD Conference on Knowledge Discovery and Data Mining},
  pages={5077--5087},
  year={2023}
}

@article{bui2022counterfactual,
  title={Counterfactual plans under distributional ambiguity},
  author={Bui, Ngoc and Nguyen, Duy and Nguyen, Viet Anh},
  journal={arXiv preprint arXiv:2201.12487},
  year={2022}
}

@article{redelmeier2021mcce,
author       = {Annabelle Redelmeier and
                Martin Jullum and
                Kjersti Aas and
                Anders L{\o}land},
title        = {{MCCE:} Monte Carlo sampling of valid and realistic counterfactual
                explanations for tabular data},
journal      = {Data Min. Knowl. Discov.},
volume       = {38},
number       = {4},
pages        = {1830--1861},
year         = {2024},
doi          = {10.1007/S10618-024-01017-Y},
}

@inproceedings{ustun2019actionable,
  title={Actionable Recourse in Linear Classification},
  author={Ustun, Berk and Spangher, Alexander and Liu, Yang},
  booktitle={Proceedings of the Conference on Fairness, Accountability, and Transparency},
  pages={10--19},
  year={2019}
}

@inproceedings{karimi2021algorithmic,
  title={Algorithmic Recourse: from Counterfactual Explanations to Interventions},
  author={Karimi, Amir-Hossein and Sch{\"o}lkopf, Bernhard and Valera, Isabel},
  booktitle={Proceedings of the 2021 ACM Conference on Fairness, Accountability, and Transparency},
  pages={353--362},
  year={2021}
}

@inproceedings{pawelczyk2020learning,
  title={Learning Model-Agnostic Counterfactual Explanations for Tabular Data},
  author={Pawelczyk, Martin and Broelemann, Klaus and Kasneci, Gjergji},
  booktitle={Proceedings of The Web Conference 2020},
  pages={3126--3132},
  year={2020}
}

@inproceedings{joshi2019towards,
  title={Towards Realistic Individual Recourse and Actionable Explanations in Black-Box Decision Making Systems},
  author={Joshi, Shalmali and Koyejo, Oluwasanmi and Vijitbenjaronk, Warut and Kim, Been and Ghosh, Joydeep},
  booktitle={NeurIPS Workshop on Human-Centric Machine Learning},
  year={2019}
}

@inproceedings{duong2023ceflow,
  title={{CeFlow}: A Robust and Efficient Counterfactual Explanation Framework for Tabular Data using Normalizing Flows},
  author={Duong, Tri and Li, Qian and Xu, Guandong},
  booktitle={Pacific-Asia Conference on Knowledge Discovery and Data Mining},
  pages={138--150},
  year={2023},
  organization={Springer}
}

@inproceedings{furman2025dicoflex,
  title={{DiCoFlex}: Model-agnostic Diverse Counterfactuals with Flexible Control},
  author={Furman, Oleksii and Movsum-zada, Ulvi and Marszalek, Patryk and Zieba, Maciej and Smieja, Marek},
  booktitle={Advances in Neural Information Processing Systems},
  volume={38},
  year={2025}
}

@inproceedings{bengio2021flow,
  title={Flow Network based Generative Models for Non-Iterative Diverse Candidate Generation},
  author={Bengio, Emmanuel and Jain, Moksh and Korablyov, Maksym and Precup, Doina and Bengio, Yoshua},
  booktitle={Advances in Neural Information Processing Systems},
  volume={34},
  pages={27381--27394},
  year={2021}
}

@inproceedings{malkin2022trajectory,
  title={Trajectory Balance: Improved Credit Assignment in {GFlowNets}},
  author={Malkin, Nikolay and Jain, Moksh and Bengio, Emmanuel and Sun, Chen and Bengio, Yoshua},
  booktitle={Advances in Neural Information Processing Systems},
  volume={35},
  pages={5955--5967},
  year={2022}
}

@inproceedings{jain2022biological,
  title={Biological Sequence Design with {GFlowNets}},
  author={Jain, Moksh and Bengio, Emmanuel and Hernandez-Garcia, Alex and Rector-Brooks, Jarrid and Dossou, Bonaventure FP and Ekbote, Chanakya and Fu, Jie and Zhang, Tianyu and Kilgour, Michael and Zhang, Dinghuai and others},
  booktitle={International Conference on Machine Learning},
  pages={9786--9801},
  year={2022},
  organization={PMLR}
}

@inproceedings{deleu2022bayesian,
  title={Bayesian Structure Learning with Generative Flow Networks},
  author={Deleu, Tristan and G{\'o}is, Ant{\'o}nio and Emezue, Chris and Rankawat, Mansi and Lacoste-Julien, Simon and Bauer, Stefan and Bengio, Yoshua},
  booktitle={Uncertainty in Artificial Intelligence},
  pages={518--528},
  year={2022},
  organization={PMLR}
}

@inproceedings{li2023dag,
  title={{DAG} Matters! {GFlowNets} Enhanced Explainer for Graph Neural Networks},
  author={Li, Wenqian and Li, Yinchuan and Li, Zhigang and Hao, Jianye and Pang, Yan},
  booktitle={International Conference on Learning Representations},
  year={2023}
}

@misc{dua2017uci,
  author       = {Hofmann, Hans},
  title        = {{Statlog (German Credit Data)}},
  year         = {1994},
  howpublished = {UCI Machine Learning Repository},
  note         = {{DOI}: https://doi.org/10.24432/C5NC77}
}

@inproceedings{kohavi1996adult,
  title={Scaling up the accuracy of naive-bayes classifiers: A decision-tree hybrid.},
  author={Kohavi, Ron and others},
  booktitle={Kdd},
  volume={96},
  pages={202--207},
  year={1996}
}

@inproceedings{acharya2019graduate,
  title={A comparison of regression models for prediction of graduate admissions},
  author={Acharya, Mohan S and Armaan, Asfia and Antony, Aneeta S},
  booktitle={2019 international conference on computational intelligence in data science (ICCIDS)},
  pages={1--5},
  year={2019},
  organization={IEEE}
}

@article{cortez2008student,
  title={Using data mining to predict secondary school student performance},
  author={Silva, Alice},
  year={2008}
}

@article{jagtiani2019lending,
  title={The roles of alternative data and machine learning in fintech lending: Evidence from the LendingClub consumer platform},
  author={Jagtiani, Julapa and Lemieux, Catharine},
  journal={Financial Management},
  volume={48},
  number={4},
  pages={1009--1029},
  year={2019},
  publisher={Wiley Online Library}
}

@article{moro2014bank,
  title={A data-driven approach to predict the success of bank telemarketing},
  author={Moro, S{\'e}rgio and Cortez, Paulo and Rita, Paulo},
  journal={Decision Support Systems},
  volume={62},
  pages={22--31},
  year={2014},
  publisher={Elsevier}
}

@article{yeh2009credit,
  title={The comparisons of data mining techniques for the predictive accuracy of probability of default of credit card clients},
  author={Yeh, I-Cheng and Lien, Che-hui},
  journal={Expert systems with applications},
  volume={36},
  number={2},
  pages={2473--2480},
  year={2009},
  publisher={Elsevier}
}

@article{goodman2017european,
  title={European Union Regulations on Algorithmic Decision-Making and a ``Right to Explanation''},
  author={Goodman, Bryce and Flaxman, Seth},
  journal={AI Magazine},
  volume={38},
  number={3},
  pages={50--57},
  year={2017}
}
\bibliographystyle{icml2026}

\newpage
\appendix
\onecolumn

\section{Diversity Guarantees}
\label{app:diversity_guarantees}
As established in the \gflownet{} literature \citep{bengio2021flow, malkin2022trajectory}, when training successfully minimizes the TB objective in Equation~\eqref{eq:tb}, the forward policy tends to sample terminal states proportionally to their reward, consistent with Equation~\eqref{eq:p_T}. In our setting, this implies:
\begin{equation}
\label{eq:reward_matching}
P_F(\mathbf{x}' | \mathbf{x}_0) \approx \frac{R(\mathbf{x}'|\mathbf{x}_0, y')}{\mathcal{Z}(\mathbf{x}_0, y')},
\end{equation}
where $\mathcal{Z}(\mathbf{x}_0, y') = \sum_{\mathbf{x}' \in \mathcal{X}} R(\mathbf{x}'|\mathbf{x}_0, y')$ is the partition function. This relationship indicates that CFs with higher reward are sampled with correspondingly higher probability.

A direct consequence concerns diversity across the counterfactual space. For any region $\mathcal{X}_A \subset \mathcal{X}$, summing Equation~\eqref{eq:reward_matching} over all $\mathbf{x}' \in \mathcal{X}_A$ yields:
\begin{equation}
    P_F(\mathcal{X}_A | \mathbf{x}_0) \approx \frac{\sum_{\mathbf{x}' \in \mathcal{X}_A}R(\mathbf{x}'|\mathbf{x}_0, y')}{\mathcal{Z}(\mathbf{x}_0, y')}. 
\end{equation}
For two disjoint regions $\mathcal{X}_A, \mathcal{X}_B \subset \mathcal{X}$, taking the ratio cancels the partition function, so the relative sampling frequencies tend toward the ratio of aggregated rewards:
\begin{equation}
\label{eq:diversity_ratio}
\frac{P_F(\mathcal{X}_A | \mathbf{x}_0)}{P_F(\mathcal{X}_B | \mathbf{x}_0)} \approx \frac{\sum_{\mathbf{x}' \in \mathcal{X}_A} R(\mathbf{x}'|\mathbf{x}_0, y')}{\sum_{\mathbf{x}' \in \mathcal{X}_B} R(\mathbf{x}'|\mathbf{x}_0, y')}.
\end{equation}
Therefore, well-trained \gflownet{} distributes probability mass across the counterfactual space in proportion to quality. This enables presenting users with diverse high-quality alternatives rather than variations of a single optimum. In practice, finite training and model capacity may yield approximate rather than exact proportionality; we evaluate this empirically in Section~\ref{sec:experiments}.

\section{Training Algorithm}
\label{apx:training_alg}

We train \our{} using a stochastic off-policy procedure outlined in Algorithm~\ref{alg:training}. The training process iterates over batches of source instances  and target labels . For each instance, the forward policy  generates a trajectory  by sequentially selecting feature indices and assigning new values until termination. Crucially, at each step , the policy is constrained by a validity mask  which enforces domain constraints (e.g., immutable features) and ensures that a feature is not modified more than once within a single trajectory.

The parameters of the forward policy  and the flow estimator  are updated jointly to minimize the Conditioned Trajectory Balance loss  (Eq.~\ref{eq:tb_loss}). As established in prior work, minimizing this objective drives the implicit sampling distribution of the policy toward the target distribution defined by the reward, . We employ a fixed, uniform backward policy , which simplifies the training dynamics by stabilizing the target flow distribution.

The reward  is computed only at terminal states, aggregating validity, sparsity, proximity, and plausibility signals. To ensure numerical stability, all computations are performed in the log domain.

\begin{algorithm}[tb]
   \caption{Training \our{} with Conditioned Trajectory Balance}
   \label{alg:training}
\begin{algorithmic}[1]
   \STATE {\bfseries Input:} Dataset $\mathcal{D}$, Classifier $h$, Reward weights $\boldsymbol{\lambda}$, Learning rate $\eta$
   \STATE {\bfseries Initialize:} Forward policy parameters $\theta$, Conditional flow parameters $\psi$
   \STATE {\bfseries Fixed:} Backward policy $P_B$ (uniform), Action mask function $\mathcal{M}$
   \REPEAT
   \STATE Sample batch of pairs $(\mathbf{x}_0, y') \sim \mathcal{D}$
   \STATE $\mathcal{L}_{\text{batch}} \gets 0$
   \FOR{each $(\mathbf{x}_0, y')$ in batch}
      \STATE Initialize state $s_0 \gets (\mathbf{x}_0, y')$, trajectory $\tau \gets \emptyset$
      \STATE $t \gets 0$
      \STATE // \textit{Rollout Trajectory}
      \WHILE{$s_t$ is not terminal}
         \STATE Get valid action mask $m_t \gets \mathcal{M}(s_t, \mathbf{x}_0)$ \COMMENT{Exclude immutable/modified features actions}
         \STATE \textit{Stage 1:} Sample feature index $d_t \sim P_F(d \mid s_t, \mathbf{x}_0, y'; \theta)$ masked by $m_t$
         \STATE \textit{Stage 2:} Sample value $v_t \sim P_F(v \mid d_t, s_t, \mathbf{x}_0, y'; \theta)$ constrained by domain limits $V(d)$
         \STATE Execute action $a_t = (d, v)$ to get $s_{t+1}$
         \STATE $\tau \gets \tau \cup \{(s_t, a_t, s_{t+1})\}$
         \STATE $t \gets t + 1$
      \ENDWHILE
      \STATE Let $\mathbf{x}' = s_t$ be the generated counterfactual
      
      \STATE // \textit{Compute Objectives}
      \STATE Compute reward $R(\mathbf{x}' | \mathbf{x}_0, y')$ using Eq.~\ref{eq:reward_composite}
      \STATE $\log P_F(\tau) \gets \sum_{i=0}^{t-1} \log P_F(a_i | s_i; \theta)$
      \STATE $\log P_B(\tau) \gets \sum_{i=0}^{t-1} \log P_B(a_i | s_{i+1})$ \COMMENT{Uniform fixed prob}
      
      \STATE // \textit{Trajectory Balance Loss (Eq.~\ref{eq:tb_loss})}
      \STATE $\Delta_{TB} \gets \log F_\psi(\mathbf{x}_0, y') + \log P_F(\tau) - \log R(\mathbf{x}'|\mathbf{x}_0,y')  - \log P_B(\tau)$
      \STATE $\mathcal{L}_{\text{batch}} \gets \mathcal{L}_{\text{batch}} + (\Delta_{TB})^2$
   \ENDFOR
   
   \STATE // \textit{Optimization Step}
   \STATE Update $\theta, \psi \gets \text{Adam}(\nabla \mathcal{L}_{\text{batch}})$
   \UNTIL{convergence}
\end{algorithmic}
\end{algorithm}

\section{Implementation Details}
\label{sec:implementation}

Our implementation builds upon the code released with \cite{koziarski2024rgfn}, which is publicly available at \href{https://github.com/koziarskilab/RGFN}{https://github.com/koziarskilab/RGFN}.


\paragraph{Architecture.}
Both policy networks, modeling actions and flow respectively, are implemented as multilayer perceptrons (MLPs) with two hidden layers of width 256 and ReLU activation function.

\paragraph{Preprocessing.}
Numerical features are discretized differently across protocols. Protocol A follows the bin ranges provided by L2C~\cite{vo2023l2c}, whereas Protocol B uses scikit-learn's \emph{KBinsDiscretizer} with a uniform discretization strategy.

\paragraph{Training.}
We optimize \our{} using the Trajectory Balance objective (Eq.~\ref{eq:tb}) with the Adam optimizer and a learning rate of $0.005$. For each dataset, training uses a batch of 1000 factual points sampled sequentially. Moreover, the reward temperature is consistently fixed at $\beta = 40.0$, and we set the \emph{logZ\_multiplier} to 10.

\paragraph{Reward Components.}
 In the discretized setting (Protocol A), we employ a simplified reward structure by setting $\lambda_v = 1$ and $\lambda_s = 0.01$, while omitting plausibility and proximity terms ($\lambda_p = \lambda_d = 0$).
In the continuous setting (Protocol B), we set the reward component weights for Adult Income, Bank, Default, and GMC datasets as follows: $\lambda_v$ (validity) = 1, $\lambda_s$ (sparsity) = 0.8, $\lambda_p$ (plausibility) = 0.4, and $\lambda_d$ (proximity) = 0.4. Furthermore, the margin parameter $\epsilon$ from the validity component is assigned to 0.1. For the Lending Club dataset, we retain the validity and plausibility weights but set $\lambda_d = 0.3$, $\lambda_s = 0.5$, and $\epsilon = 0.25$. For all datasets, the proximity component is computed over numerical features, and sparsity over categorical ones.

\paragraph{Generation.}
For each test instance, we sample $K=10$ CFs via forward policy rollouts. All experiments in Protocol A are repeated over 5 random seeds, while those in Protocol B use a single seed.

\paragraph{Hardware.}
All training and evaluation were performed on a single NVIDIA GeForce RTX 4090 GPU. Each dataset required less than 10 hours of training.

\section{Baselines}
\label{app:baselines}

We compare \our{} against a comprehensive set of state-of-the-art methods, spanning three major paradigms in counterfactual explanation generation.

\paragraph{Optimization-based methods} generate CFs by optimizing an objective function for each input instance individually.
\begin{compactitem}
    \item \textbf{DiCE} \citep{mothilal2020dice} is a prominent method that generates a diverse set of CFs by adding a Determinantal Point Process (DPP) term to the optimization objective. It handles both continuous and categorical features. For Protocol A, we use the random search (\textbf{DICE-R}) and genetic algorithm (\textbf{DICE-G}) variants. For Protocol B, we use the KD-Tree-based search.
    \item \textbf{COPA} \citep{bui2022counterfactual} generates CFs for structured policies and has been adapted for tabular data. It is included in Protocol A.
\end{compactitem}

\paragraph{Sampling-based methods} explore the feature space to find CFs.
\begin{compactitem}
    \item \textbf{MCCE} \citep{redelmeier2021mcce} builds a local model around the instance to be explained and then uses Monte Carlo sampling from a conditional distribution of mutable features to generate CFs that lie on the data manifold. It is evaluated under Protocol A.
\end{compactitem}

\paragraph{Generative and Amortized methods} train a single model to generate CFs for any input instance in a single forward pass, amortizing the computational cost.
\begin{compactitem}
    \item \textbf{L2C} \citep{vo2023l2c} is an amortized framework that learns feature-based perturbation and selection distributions to generate diverse and privacy-preserving CFs on discretized data. It is a key baseline in Protocol A.
    \item \textbf{C-CHVAE} \citep{pawelczyk2020learning} uses a Conditional Variational Autoencoder (CVAE) to learn class-conditional latent representations, generating plausible CFs by searching in the latent space. It is evaluated under Protocol B.
    \item \textbf{DiCoFlex} \citep{furman2025dicoflex} employs a conditional normalizing flow to generate diverse CFs in a single forward pass. It allows for inference-time control over sparsity and actionability constraints. It is a key baseline in Protocol B.
\end{compactitem}

\section{Evaluation datasets}
\label{app:datasets}

\begin{table}[t]
\caption{Dataset characteristics. Protocol A: discretized ($B=4$); Protocol B: continuous with generation-time quantization ($B=64$).}
\label{tab:datasets}
\centering
\begin{tabular}{lccccc}
\toprule
\textbf{Dataset} & \textbf{Protocol} & $n$ & $d_{\text{cont}}$ &
$d_{\text{cat}}$ & \textbf{Imbalance} \\
\midrule
German Credit & A & 1,000 & 3 & 17 & 30\% \\
Adult & A,B & 32,561 & 4 & 8 & 24\% \\
Graduate Adm. & A & 500 & 3 & 4 & 34\% \\
Student Perf. & A & 649 & 4 & 10 & 33\% \\
\midrule
Lending Club & B & 30,000 & 8 & 4 & 22\% \\
GMC & B & 30,000 & 6 & 3 & 7\% \\
Bank Marketing & B & 30,000 & 7 & 9 & 11\% \\
Default & B & 27,000 & 14 & 9 & 22\% \\
\bottomrule
\end{tabular}
\end{table}

In this appendix, we detail the datasets used to train and evaluate \our{}. Table~\ref{tab:datasets} summarizes their key characteristics, including the number of continuous ($d_{\text{cont}}$) and categorical ($d_{\text{cat}}$) features, the target class imbalance, and the assigned evaluation protocol. In the following subsections, we provide a description for each dataset individually.

\subsection{Protocol A Datasets}

\textbf{German Credit.}
The German Credit dataset~\citep{dua2017uci} from the UCI Machine Learning Repository contains information about bank customers applying for credit. Each instance includes 20 attributes covering demographic information (age, personal status, housing), financial indicators (credit amount, savings, checking account status), and employment details. The binary classification task is to assess whether a customer presents a good or bad credit risk. The dataset comprises 1,000 instances with a 30\% minority class ratio. Available at: \url{https://archive.ics.uci.edu/ml/datasets/statlog+(german+credit+data)}.

\textbf{Adult Income.}
The Adult Census Income dataset~\citep{kohavi1996adult}, extracted from the 1994 U.S.\ Census database, contains demographic information such as age, education, occupation, work hours per week, and capital gain/loss. The binary classification task is to predict whether an individual's annual income exceeds \$50,000. With 32,561 samples and a 24\% positive class rate, this dataset is widely used in fairness and explainability research due to the presence of sensitive attributes (race, gender, age). This dataset is used under both Protocol A and Protocol B. Available at: \url{https://archive.ics.uci.edu/ml/datasets/adult}.

\textbf{Graduate Admission.}
The Graduate Admission dataset~\citep{acharya2019graduate} contains records of Indian students' applications to Master's programs, including GRE scores, TOEFL scores, university rating, statement of purpose strength, letter of recommendation strength, undergraduate GPA, and research experience. The original target is a continuous admission probability; following~\citet{vo2023l2c}, we binarize it at a threshold of 0.7 to classify students as having a higher or lower chance of admission. The dataset contains 500 instances. Available at: \url{https://www.kaggle.com/datasets/mohansacharya/graduate-admissions}.

\textbf{Student Performance.}
The Student Performance dataset~\citep{cortez2008student} records academic outcomes of secondary school students from two Portuguese schools, with 14 features including demographic attributes (age, family size), social factors (parents' education, family support), and academic indicators (study time, first-period grades). The binary classification task is to predict whether a student achieves a final score above average. The dataset contains 649 instances with train/test splits corresponding to the two schools. Available at: \url{https://archive.ics.uci.edu/ml/datasets/student+performance}.

\subsection{Protocol B Datasets}

\textbf{Lending Club.}
The Lending Club dataset~\citep{jagtiani2019lending} contains detailed information about loans issued through the Lending Club peer-to-peer lending platform. Features include borrower characteristics (credit score, annual income, employment length), loan specifics (loan amount, interest rate, purpose), and performance indicators (payment status, delinquency). The binary classification task is to predict whether a loan will be fully paid or charged off (default). The dataset contains 30,000 samples with a 22\% default rate. Available at: \url{https://www.kaggle.com/datasets/wordsforthewise/lending-club}.

\textbf{Give Me Some Credit (GMC).}
The Give Me Some Credit dataset~\citep{pawelczyk2020learning} contains anonymized records of credit users with features such as debt-to-income ratio, number of delinquencies, monthly income, age, and number of open credit lines. The target variable indicates whether a user experienced a serious delinquency (more than 90 days overdue) within the previous two years. The dataset contains 30,000 samples with a highly imbalanced 7\% positive class rate. Available at: \url{https://www.kaggle.com/c/GiveMeSomeCredit}.

\textbf{Bank Marketing.}
The Bank Marketing dataset~\citep{moro2014bank} contains information from a direct marketing campaign conducted by a Portuguese banking institution. Features include client data (age, job, marital status, education), campaign contact information (communication type, day, month), economic indicators, and previous campaign outcomes. The prediction task is to determine whether a client will subscribe to a term deposit. The dataset comprises 30,000 samples with an 11\% positive class rate. Available at: \url{https://archive.ics.uci.edu/ml/datasets/bank+marketing}.

\textbf{Credit Default.}
The Credit Default dataset~\citep{yeh2009credit} contains information on credit card clients in Taiwan, including demographic factors, credit data, payment history, and bill statements. The target variable indicates whether the client defaulted on their payment in the following month. With 23 features (14 numerical and 9 categorical) and 27,000 samples, this dataset presents complex feature interdependencies common in financial data. Available at: \url{https://archive.ics.uci.edu/ml/datasets/default+of+credit+card+clients}.

\section{Evaluation metrics}
\label{app:metrics}

This section provides detailed descriptions of the evaluation metrics used in our experiments, organized by protocol. The two protocols are adopted from two different works, \citet{vo2023l2c} (Protocol A) and \citet{furman2025dicoflex} (Protocol B). The main difference between them is that Protocol A assumes all features are discretized, while Protocol B works with both continuous and categorical features. Furthermore, Protocol A reports metrics normalized to percentages. Table~\ref{tab:metrics_protocol_a} outlines the metrics for Protocol A, while Table~\ref{tab:metrics_protocol_b} details the metrics for Protocol B.

\begin{table}[h]
\caption{Description of quantitative evaluation metrics for Protocol A. For each test instance $i \in \{1,\dots,N\}$, we generate a set of $K$ counterfactuals $\mathcal{C}_i = \{\mathbf{x}'^{(k)}_i\}_{k=1}^K$. Let $h$ denote the classifier, $y'_i$ the desired target class, $D$ the total number of features, and $\mathcal{F}_m$ the set of features subject to monotonic constraints.}
\label{tab:metrics_protocol_a}
\centering
\begin{tabular}{llp{3.8cm}p{8cm}}
\toprule
\textbf{Desideratum} & \textbf{Metric} & \textbf{Description} & \textbf{Formula} \\
\midrule
\multirow{2}{*}{Validity ($\uparrow$)} & Val.(\%) & Proportion of valid counterfactuals. & $\displaystyle\frac{1}{N \cdot K} \sum_{i=1}^{N} \sum_{k=1}^{K} \mathbb{1}\!\left[h(\mathbf{x}'^{(k)}_i) = y'_i\right]$ \\[6pt]
& Cov.(\%) & Proportion of instances with at least one valid CF. & $\displaystyle\frac{1}{N} \sum_{i=1}^{N} \mathbb{1}\!\left[\exists\, k : h(\mathbf{x}'^{(k)}_i) = y'_i\right]$ \\[6pt]
\midrule
Sparsity ($\uparrow$) & Spars.(\%) & Proportion of features that remain unchanged. & $\displaystyle\frac{1}{N \cdot K} \sum_{i=1}^{N} \sum_{k=1}^{K} \frac{1}{D}\sum_{j=1}^{D} \mathbb{1}\!\left[x'^{(k)}_{i,j} = x_{i,j}\right]$ \\[6pt]
\midrule
Diversity ($\uparrow$) & Div.(\%) & Mean pairwise Hamming distance. & $\displaystyle\frac{1}{N}\sum_{i=1}^{N}\frac{2}{K(K\!-\!1)}\!\sum_{k<l}\frac{1}{D}\sum_{j=1}^{D}\mathbb{1}\!\left[x'^{(k)}_{i,j} \neq x'^{(l)}_{i,j}\right]$ \\[6pt]
\midrule
Balance ($\uparrow$) & H. Mean(\%) & Harmonic mean of Diversity and Sparsity. & $\displaystyle\frac{2 \cdot \text{Div} \cdot \text{Spars.}} {\text{Div} + \text{Spars.}}$ \\[6pt]
\midrule
Constraints ($\uparrow$) & Unary(\%) & Proportion of CFs satisfying monotonic constraints. & $\displaystyle\frac{1}{N \cdot K}\sum_{i=1}^{N}\sum_{k=1}^{K}\frac{1}{|\mathcal{F}_m|}\sum_{j \in \mathcal{F}_m}\mathbb{1}\!\left[\text{mon}_j(\mathbf{x}'^{(k)}_i, \mathbf{x}_i)\right]$ \\
\bottomrule
\end{tabular}
\end{table}

\begin{table}[h]
\caption{Description of quantitative evaluation metrics for Protocol B. We generate $N$ counterfactuals $\{\mathbf{x}'_n\}_{n=1}^N$ for a test set of $N$ instances $\{\mathbf{x}^0_n\}_{n=1}^N$, of which $N_{\text{val}}$ are valid. Let $h$ denote the classifier, $D_{\text{cat}}$ and $D_{\text{num}}$ the number of categorical and continuous features, $D=D_{\text{cat}}+D_{\text{num}}$ the total, $V_j$ the range of feature $j$, $G$ the number of unique original instances (groups), $K_g$ the number of counterfactuals for group $g$, and $\mathcal{N}$, $\mathcal{C}$ the sets of numerical and categorical feature indices.}
\label{tab:metrics_protocol_b}
\centering
\begin{tabular}{llp{3.5cm}p{8cm}}
\toprule
\textbf{Desideratum} & \textbf{Metric} & \textbf{Description} & \textbf{Formula} \\
\midrule
Validity ($\uparrow$) & Val. & Success rate of changing model predictions. & $\displaystyle\frac{1}{N}\sum_{n=1}^{N}\mathbb{1}\!\left[h(\mathbf{x}'_n) = y'\right]$ \\[6pt]
\midrule
Proximity ($\downarrow$) & Prox.-Cont & $L_1$ distance on continuous features. & $\displaystyle\frac{1}{N_{\text{val}}}\sum_{n=1}^{N_{\text{val}}}\left\|\mathbf{x}^0_{n,\text{num}} - \mathbf{x}'^{\,}_{n,\text{num}}\right\|_1$ \\[6pt]
\midrule
\multirow{2}{*}{Sparsity ($\downarrow$)}
    & Spars.-Cat & Proportion of changed categorical features. & $\displaystyle\frac{1}{N_{\text{val}}}\sum_{n=1}^{N_{\text{val}}}\frac{\left\|\mathbf{x}^0_{n,\text{cat}} - \mathbf{x}'^{\,}_{n,\text{cat}}\right\|_0}{D_{\text{cat}}}$ \\[6pt]
    & $\epsilon$-Spars. & Continuous features changed beyond $\epsilon \cdot V_j$ ($\epsilon{=}0.05$). & $\displaystyle\frac{1}{N_{\text{val}}}\sum_{n=1}^{N_{\text{val}}}\frac{1}{D_{\text{num}}}\sum_{j=1}^{D_{\text{num}}}\mathbb{1}\!\left[|x^0_{n,j} - x'_{n,j}| > \epsilon \cdot V_j\right]$ \\[6pt]
\midrule
Plausibility ($\downarrow$) & LOF & Median Local Outlier Factor (log scale). & $\displaystyle\text{median}_{n=1}^{N_{\text{val}}}\;\log\!\left(\text{LOF}_k(\mathbf{x}'_n)\right)$ \\[6pt]
\midrule
Diversity ($\uparrow$) & Div. & Mean pairwise mixed distance between counterfactuals, averaged over groups. & $\displaystyle\frac{1}{G}\sum_{g=1}^{G}\frac{1}{D\binom{K_g}{2}}\sum_{i<j}^{K_g}\!\left(\left\|\mathbf{x}'_{i,\text{num}} - \mathbf{x}'^{\,}_{j,\text{num}}\right\|_1 + \left\|\mathbf{x}'_{i,\text{cat}} - \mathbf{x}'^{\,}_{j,\text{cat}}\right\|_0\right)$ \\
\bottomrule
\end{tabular}
\end{table}

\begin{figure*}[t]
  \centering
  \begin{subfigure}[b]{0.32\textwidth}
    \centering
    \includegraphics[width=\textwidth]{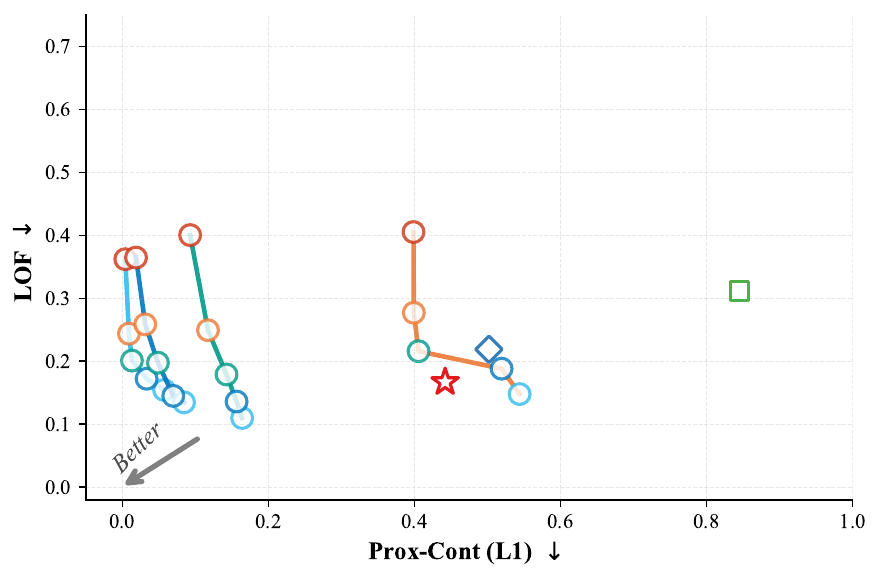}
    \caption{Proximity vs Plausibility}
    \label{fig:ablation-a-all}
  \end{subfigure}
  \hfill
  \begin{subfigure}[b]{0.32\textwidth}
    \centering
    \includegraphics[width=\textwidth]{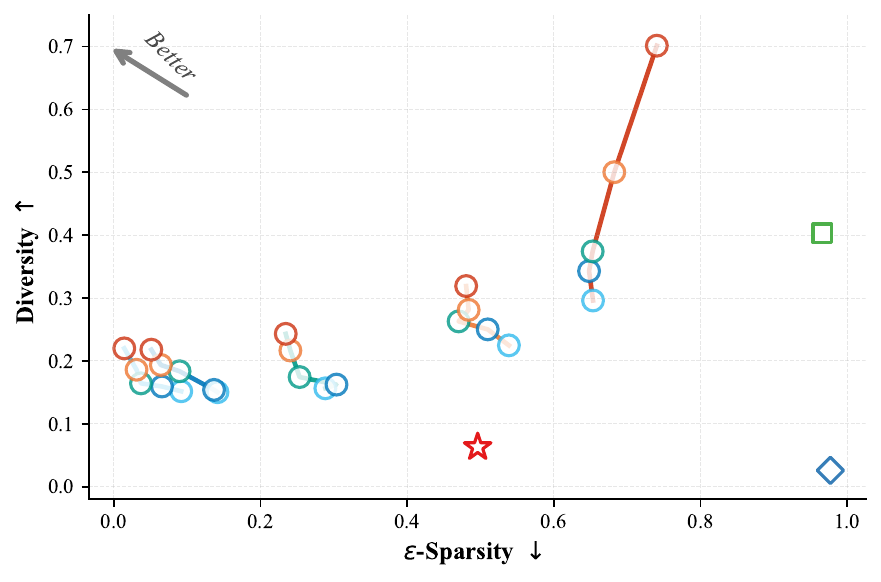}
    \caption{Diversity vs Sparsity}
    \label{fig:ablation-c-all}
  \end{subfigure}
  \hfill
  \begin{subfigure}[b]{0.32\textwidth}
    \centering
    \includegraphics[width=\textwidth]{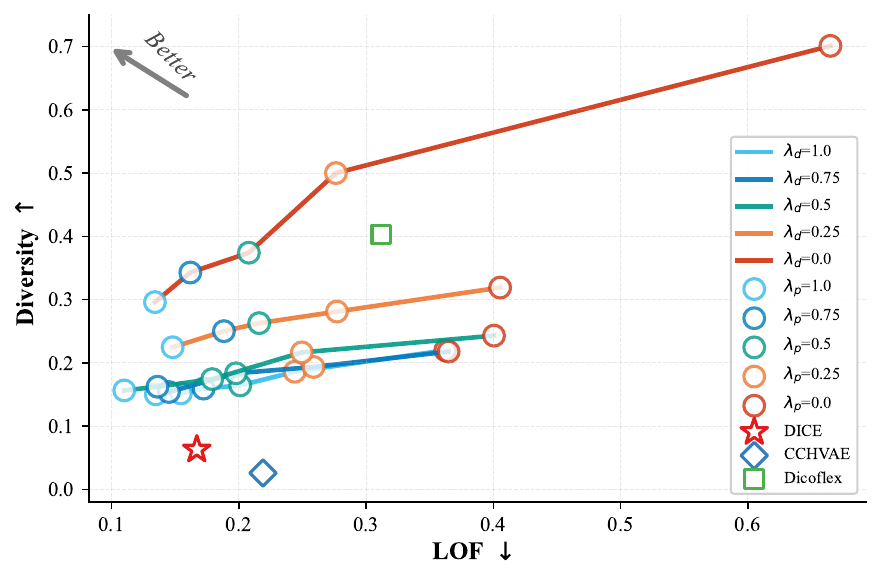}
    \caption{Diversity vs Plausibility}
    \label{fig:ablation-b-all}
  \end{subfigure}
  \caption{Ablation of $\lambda_p$ and $\lambda_d$ with fixed weights for other reward weights ($\lambda_v=1.0,\lambda_s=0.4$)}
  \label{fig:ablation_weights_all}
\end{figure*}

\section{Reward Components Ablation}
\label{app:reward_ablation}

We perform an extensive ablation study to analyze the effect of the reward component weights on the generated CFs. The composite reward (Eq.~\ref{eq:reward_composite}) is controlled by four weights: $\lambda_v$ (validity), $\lambda_d$ (proximity), $\lambda_p$ (plausibility), and $\lambda_s$ (sparsity). In our experiments, we keep $\lambda_v=1.0$ and $\lambda_s=0.8$ fixed and perform a grid search over $\lambda_d$ and $\lambda_p$ on the Adult dataset.

The results are visualized in Figure~\ref{fig:ablation_weights_all}. The quantitative results confirm the trade-offs discussed in the main paper. Specifically, we observe that decreasing the proximity weight ($\lambda_d$) consistently leads to an increase in diversity, but also a worsening of proximity (L1), sparsity ($\epsilon$-sparsity), and plausibility (LOF). Similarly, decreasing the plausibility weight ($\lambda_p$) improves proximity and sparsity at the expense of plausibility. For instance, with $\lambda_d=1.0$, decreasing $\lambda_p$ from 1.0 to 0.0 reduces the L1 distance, but increases the LOF score.

\section{Feature Utilization}
\label{app:feature_utilization}

\begin{figure}[th]
\centering
\begin{subfigure}[b]{0.45\textwidth}
    \centering
    \includegraphics[width=\textwidth]{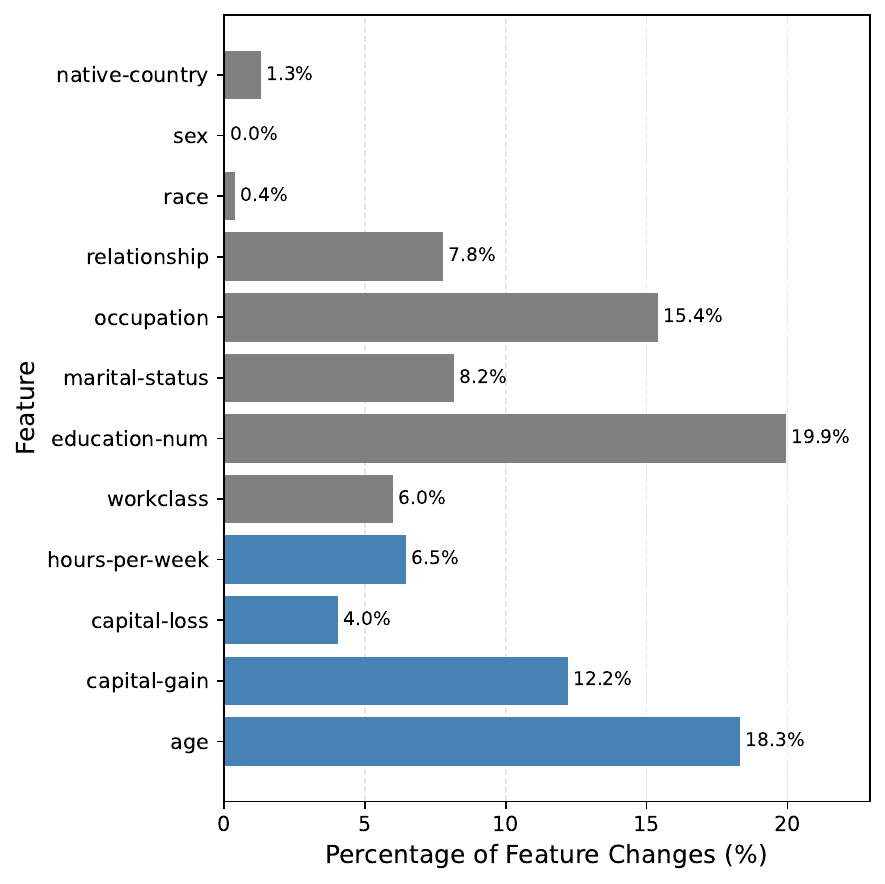}
    \caption{\our{}}
    \label{fig:feature_utilization_our}
\end{subfigure}
\begin{subfigure}[b]{0.45\textwidth}
    \centering
    \includegraphics[width=\textwidth]{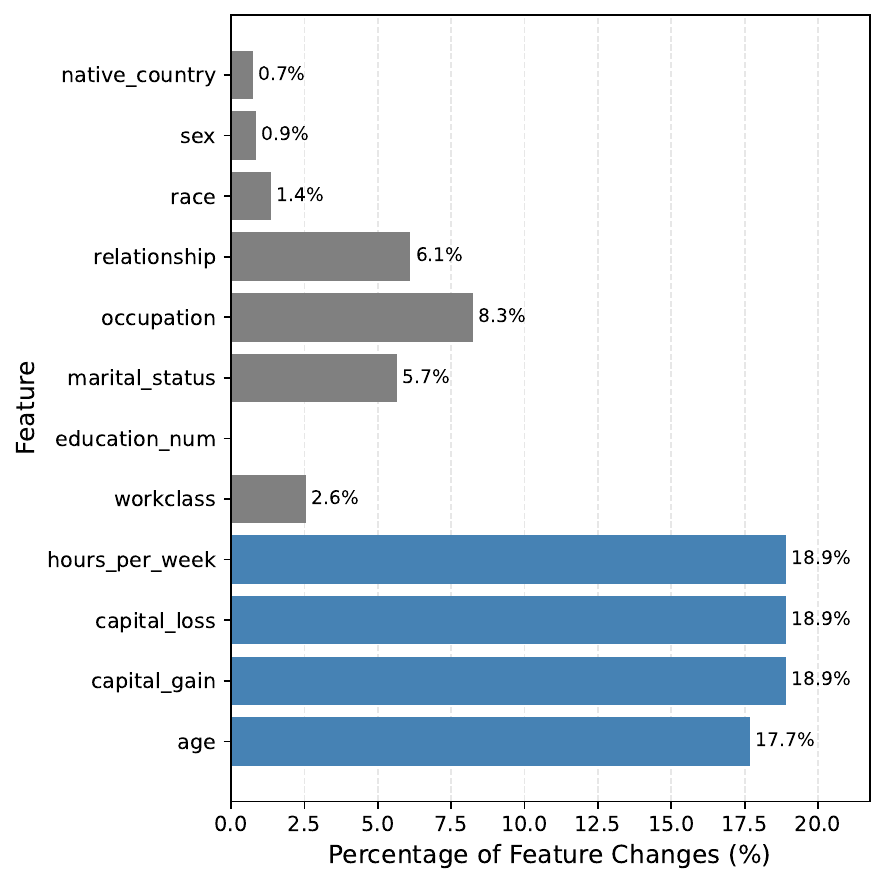}
    \caption{DiCE}
    \label{fig:feature_utilization_dice}
\end{subfigure}
\begin{subfigure}[b]{0.45\textwidth}
    \centering
    \includegraphics[width=\textwidth]{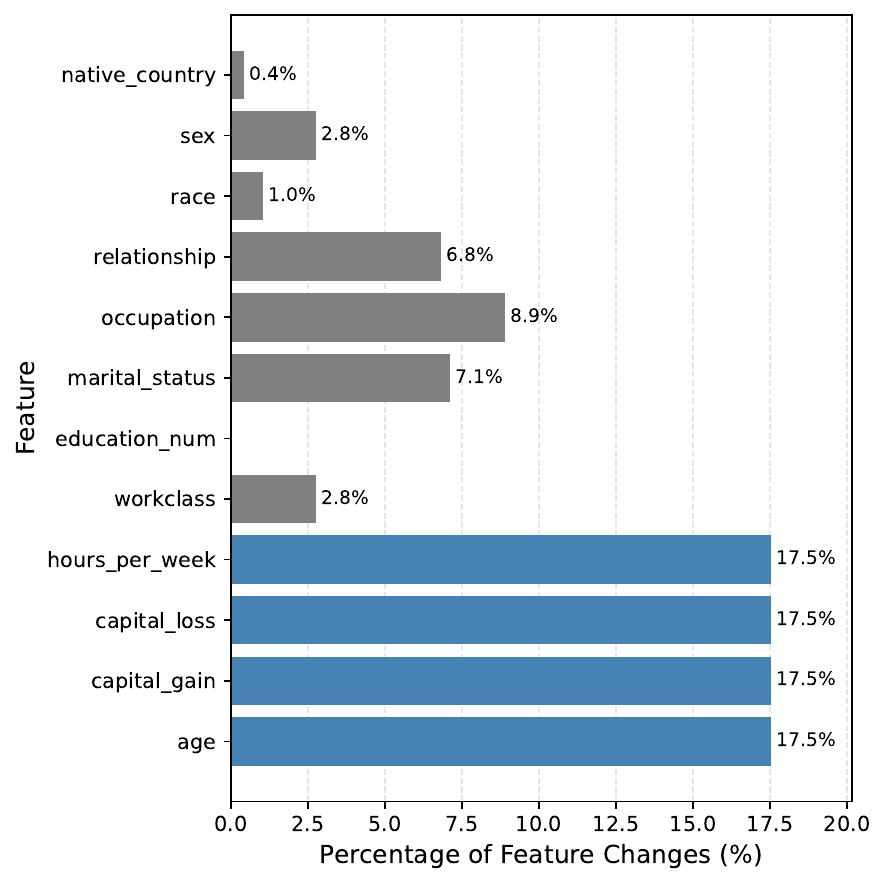}
    \caption{CCHVAE}
    \label{fig:feature_utilization_cchvae}
\end{subfigure}
\begin{subfigure}[b]{0.45\textwidth}
    \centering
    \includegraphics[width=\textwidth]{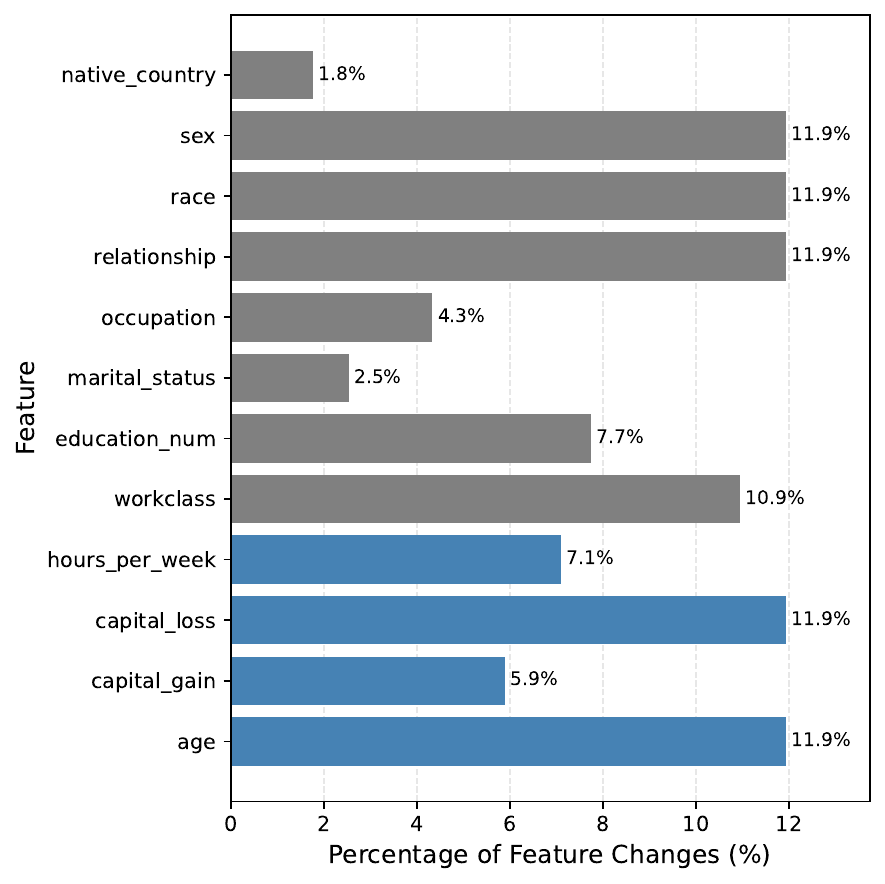}
    \caption{DiCoFlex}
    \label{fig:feature_utilization_dicoflex}
\end{subfigure}
\caption{Visualization of feature utilization of \our{} method and other baseline methods on the Adult dataset. The plots show the frequency with which each feature is modified in the generated CFs. \our{} demonstrates a more balanced feature utilization compared to baselines.}
\label{fig:feature_utilization}
\end{figure}

We analyze the feature utilization of \our{} and baseline methods on the Adult dataset to understand which features are most frequently modified to generate CFs. \Cref{fig:feature_utilization} illustrates the percentage of changes for each feature. \our{} (\Cref{fig:feature_utilization_our} demonstrates a balanced utilization of both numerical and categorical features. The most frequently modified features include `education-num` (19.9\%), `age` (18.3\%), and `occupation` (15.4\%), indicating that \our{} leverages a diverse set of attributes to find recourse. In contrast, DiCE (\Cref{fig:feature_utilization_dice} and CCHVAE (\Cref{fig:feature_utilization_cchvae} exhibit a strong bias towards numerical features, with changes heavily concentrated on `age`, `capital-gain`, `capital-loss`, and `hours-per-week` (all above 17\%). This suggests that these methods may overlook viable CFs that involve changes to categorical attributes. DiCoFlex (\Cref{fig:feature_utilization_dicoflex} shows a more uniform, yet still skewed, distribution, with a wider range of features being modified but still a noticeable preference for certain features over others.

\section{Qualitative Examples}
\label{app:qualitative_examples}

This section provides qualitative examples of CFs generated by \our{} for the Adult (\Cref{tab:qualitative_adult}) and GMC (\Cref{tab:qualitative_gmc}) datasets.    For each dataset, we present two instances, showing the original input and two diverse, valid CFs that change the prediction from low income to high income (Adult) or from high risk to low risk (GMC). Changed features are highlighted in         \textbf{bold}.

\begin{table}[h!]
\centering
\caption{Counterfactual examples for the Adult dataset. The original prediction is "low income," and the counterfactual          prediction is "high income."}
\label{tab:qualitative_adult}
\begin{tabular}{l|c|cc}
\toprule
\textbf{Feature} & \textbf{Original Input} & \textbf{Counterfactual 1} & \textbf{Counterfactual 2} \\
\midrule
Age & 25 & 25 & \textbf{29} \\
Workclass & Self-emp-not-inc & \textbf{Federal-gov} & Self-emp-not-inc \\
Education & HS-grad & \textbf{Prof-school} & HS-grad \\
Marital Status & Never-married & \textbf{Married-AF-spouse} & \textbf{Married-civ-spouse} \\
Occupation & Sales & Sales & Sales \\
Relationship & Other-relative & \textbf{Wife} & \textbf{Wife} \\
Race & Asian-Pac-Islander & Asian-Pac-Islander & Asian-Pac-Islander \\
Sex & Male & Male & Male \\
Capital Gain & 781 & 781 & \textbf{7031} \\
Capital Loss & 34 & 34 & 34 \\
Hours per Week & 50 & 50 & 50 \\
Native Country & United-States & United-States & United-States \\
\bottomrule
\end{tabular}
\end{table}

\begin{table}[h!]
\centering
\caption{Counterfactual examples for the GMC dataset. The original prediction is "high risk," and the counterfactual prediction is "low risk."}
\label{tab:qualitative_gmc}
\begin{tabular}{l|c|cc}
\toprule
\textbf{Feature} & \textbf{Original Input} & \textbf{Counterfactual 1} & \textbf{Counterfactual 2} \\
\midrule
Revolving Util. of Unsecured Lines & 69 & 69 & 69 \\
Age & 45 & \textbf{21} & \textbf{56} \\
Times 30-59 Days Past Due & 1 & \textbf{0} & \textbf{3} \\
Debt Ratio & 162.57 & 162.57 & 162.57 \\
Monthly Income & 5488.28 & 5488.28 & 5488.28 \\
Open Credit Lines \& Loans & 5 & \textbf{0} & \textbf{4} \\
Times 90+ Days Late & 0 & 0 & 0 \\
Real Estate Loans/Lines & 0 & 0 & 0 \\
Times 60-89 Days Past Due & 0 & 0 & \textbf{3} \\
Number of Dependents & 0 & 0 & 0 \\
\bottomrule
\end{tabular}
\end{table}

\end{document}